\documentclass[pdflatex,sn-nature]{sn-jnl}

\usepackage{graphicx}
\usepackage{placeins}
\usepackage{multirow}
\usepackage{amsmath,amssymb,amsfonts}
\usepackage{xcolor}
\usepackage{textcomp}
\usepackage{booktabs}
\usepackage{array}
\usepackage{tabularx}
\usepackage{caption}
\usepackage{float}
\usepackage{helvet}
\usepackage{pdflscape}
\usepackage{makecell}
\usepackage{url}
\usepackage{hyperref}
\graphicspath{{figures/}}

\newif\ifshowqueries
\showqueriesfalse
\ifdefined\CLEANCOPY\showqueriesfalse\fi
\DeclareRobustCommand{\AQ}[1]{\ifshowqueries\textsuperscript{\textcolor{red}{\textbf{[AQ#1]}}}\fi}

\newcommand{\EDtabfont}{\fontfamily{phv}\selectfont}
\newcolumntype{L}{>{\EDtabfont\raggedright\arraybackslash}l}
\newcolumntype{C}{>{\EDtabfont\centering\arraybackslash}c}
\newcolumntype{Y}[1]{>{\EDtabfont\raggedright\arraybackslash}p{#1}}

\newif\iffiguresinline
\figuresinlinetrue
\ifdefined\FIGURESATEND\figuresinlinefalse\fi

\renewcommand\Authorfont{\normalfont\normalsize\boldmath\titraggedcenter}
\renewcommand\addressfont{\normalfont\normalsize\titraggedcenter}

\usepackage{lineno}

\newcommand*\patchAmsMathEnvironmentForLineno[1]{%
  \expandafter\let\csname old#1\expandafter\endcsname\csname #1\endcsname
  \expandafter\let\csname oldend#1\expandafter\endcsname\csname end#1\endcsname
  \renewenvironment{#1}%
     {\linenomath\csname old#1\endcsname}%
     {\csname oldend#1\endcsname\endlinenomath}}
\AtBeginDocument{\patchAmsMathEnvironmentForLineno{equation}}

\newcommand{\figI}[2]{%
\begin{figure}[#1]
\centering
\includegraphics[width=#2]{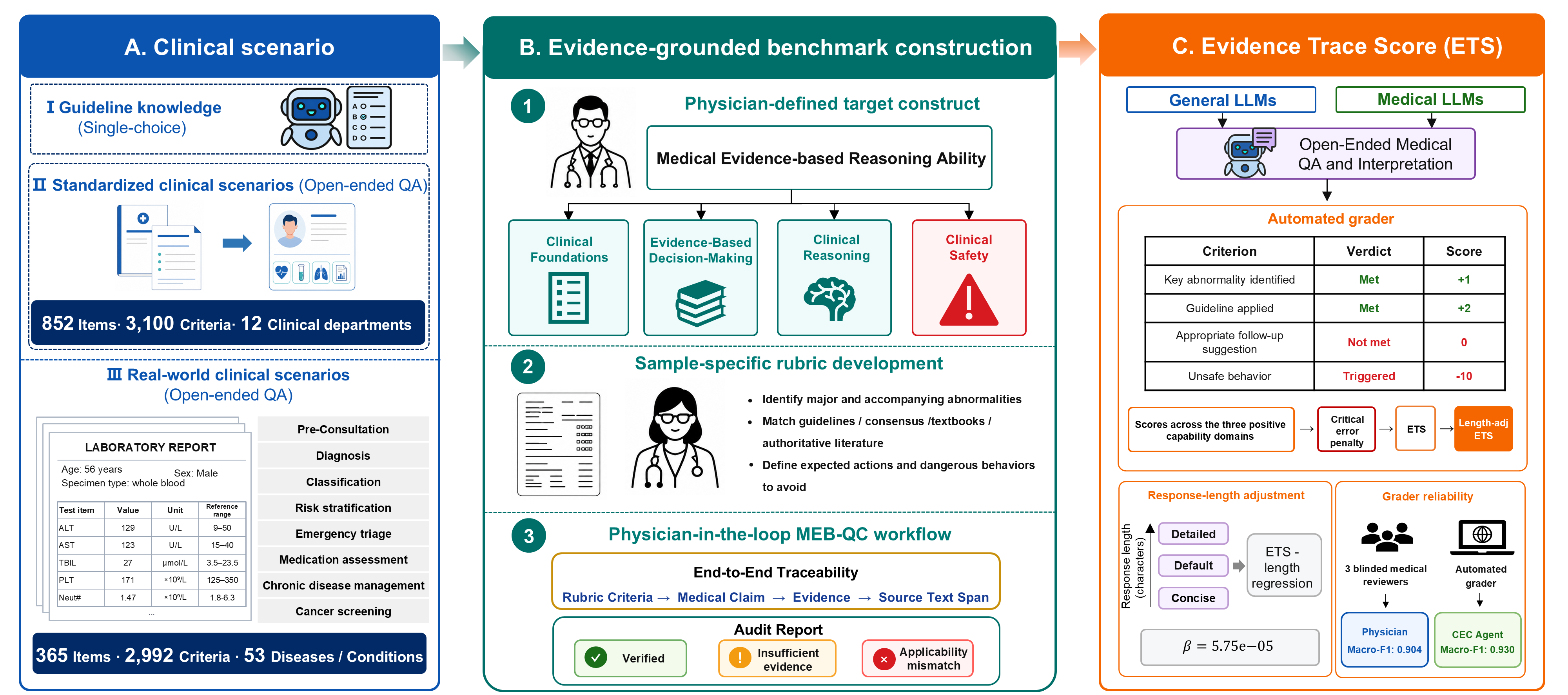}
\caption{\textbf{Construction and evaluation framework of MEB.} \textbf{a}, Evaluation levels. MEB comprises 319 guideline-knowledge items (single-best-answer questions), 852 standardized clinical scenarios (3,100 criteria) and 365 real-world clinical scenarios built from laboratory records (2,992 criteria), classified by eight clinical intents. \textbf{b}, Construction of the benchmark. Physicians defined the target construct and its four capability domains, wrote item-specific rubrics, and verified each criterion through the physician-in-the-loop MEB-QC workflow, which links each criterion to a medical claim, an evidence record and a source-text span. \textbf{c}, Scoring. The CEC Agent adjudicated each criterion independently; criterion-level judgments were aggregated into the Evidence Trace Score (ETS), which was examined for sensitivity to response length, and the grader was compared with three blinded physician reviewers.}
\label{fig:1}
\end{figure}}
\newcommand{\figII}[2]{%
\begin{figure}[#1]
\centering
\includegraphics[width=#2]{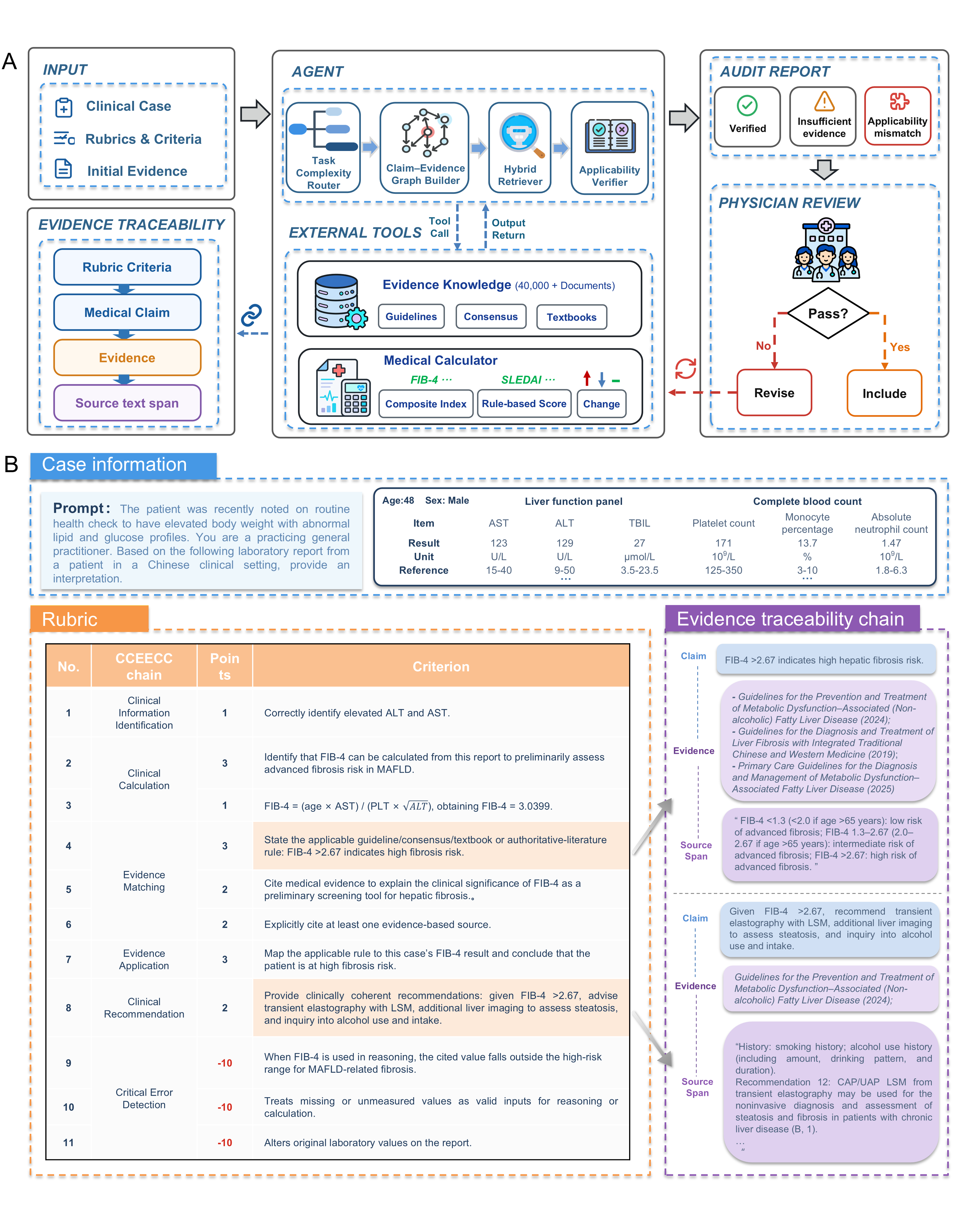}
\caption{\textbf{Quality control of item-specific rubrics and an example of evidence traceability.} \textbf{a}, The MEB-QC workflow. Physicians submitted a case, its rubric and the initial supporting evidence. The MEB-QC Agent audited each criterion against the controlled evidence corpus and returned a structured report that classified the criterion as verified, insufficient evidence or applicability mismatch. Physicians reviewed each finding, and revised rubrics re-entered the audit until they were accepted for inclusion. \textbf{b}, Example of a real-world scenario concerning risk stratification of liver fibrosis in metabolic dysfunction-associated fatty liver disease (MAFLD), showing the prompt in English translation  and an excerpt of the laboratory report, the criteria organized by CCEECC stage with their weights, and the chain from criterion to medical claim, evidence and source-text span for two criteria (shaded). ALT, alanine aminotransferase; AST, aspartate aminotransferase; CAP/UAP, controlled attenuation parameter/ultrasound attenuation parameter; FIB-4, fibrosis-4 index; LSM, liver stiffness measurement; SLEDAI, systemic lupus erythematosus disease activity index; TBIL, total bilirubin.}
\label{fig:2}
\end{figure}}
\newcommand{\figIII}[2]{%
\begin{figure}[#1]
\centering
\includegraphics[width=#2]{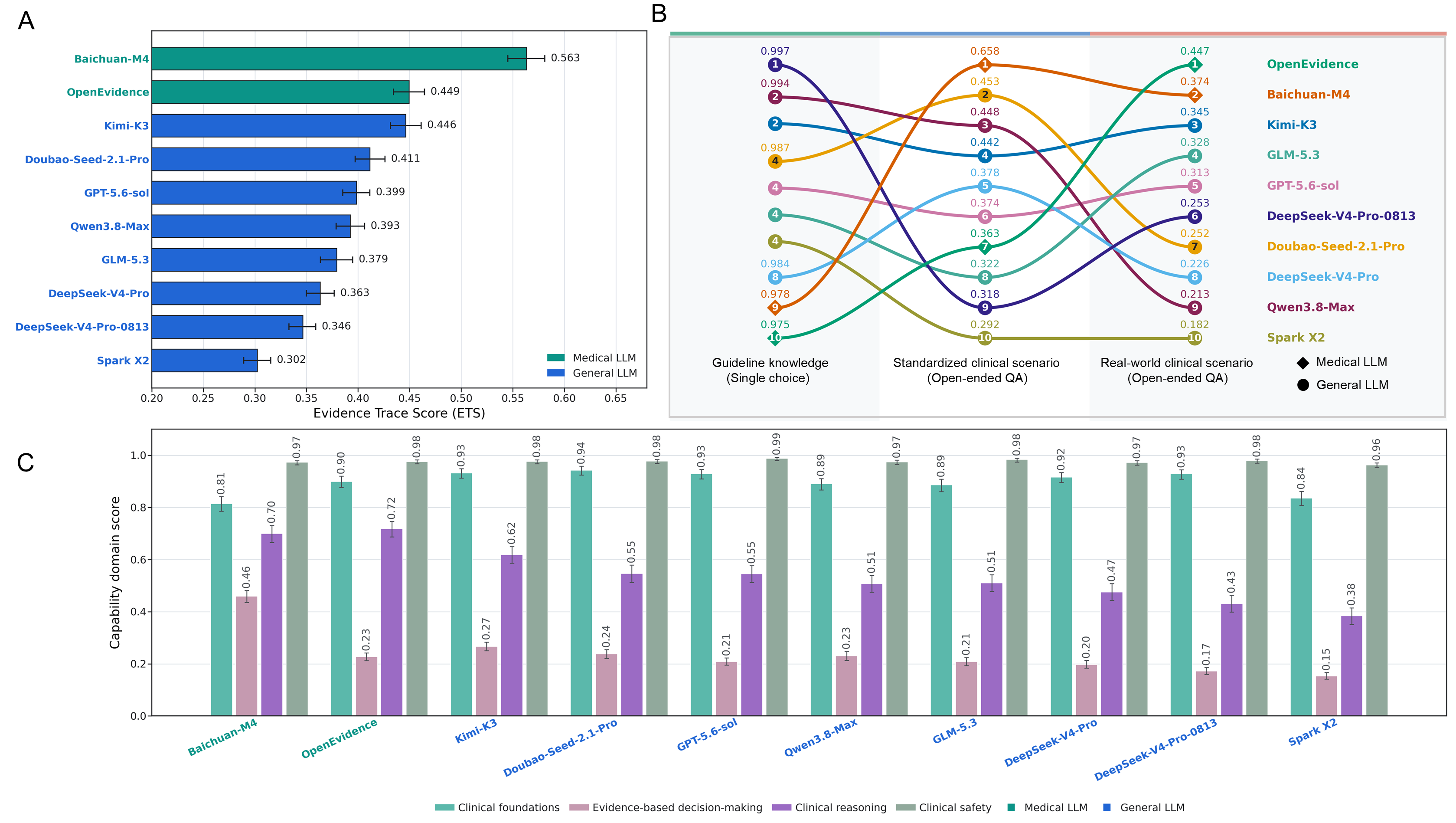}
\caption{\textbf{Performance of ten LLMs on MEB.} \textbf{a}, Overall ETS across all 1,536 items. Bars show point estimates and error bars show 95\% CIs from 5,000 case-level bootstrap resamples stratified by evaluation level. \textbf{b}, ETS (values above the markers) and rank (numbers within the markers) of each model for guideline knowledge ($n = 319$ items), standardized clinical scenarios ($n = 852$) and real-world clinical scenarios ($n = 365$). Diamonds denote medical systems and circles denote general-purpose models; models with equal scores share a rank. \textbf{c}, Scores in the four capability domains across all items; error bars are defined as in \textbf{a}. In \textbf{a} and \textbf{c}, model names in green denote medical systems and those in blue denote general-purpose models.}
\label{fig:3}
\end{figure}}
\newcommand{\figIV}[2]{%
\begin{figure}[#1]
\centering
\includegraphics[width=#2]{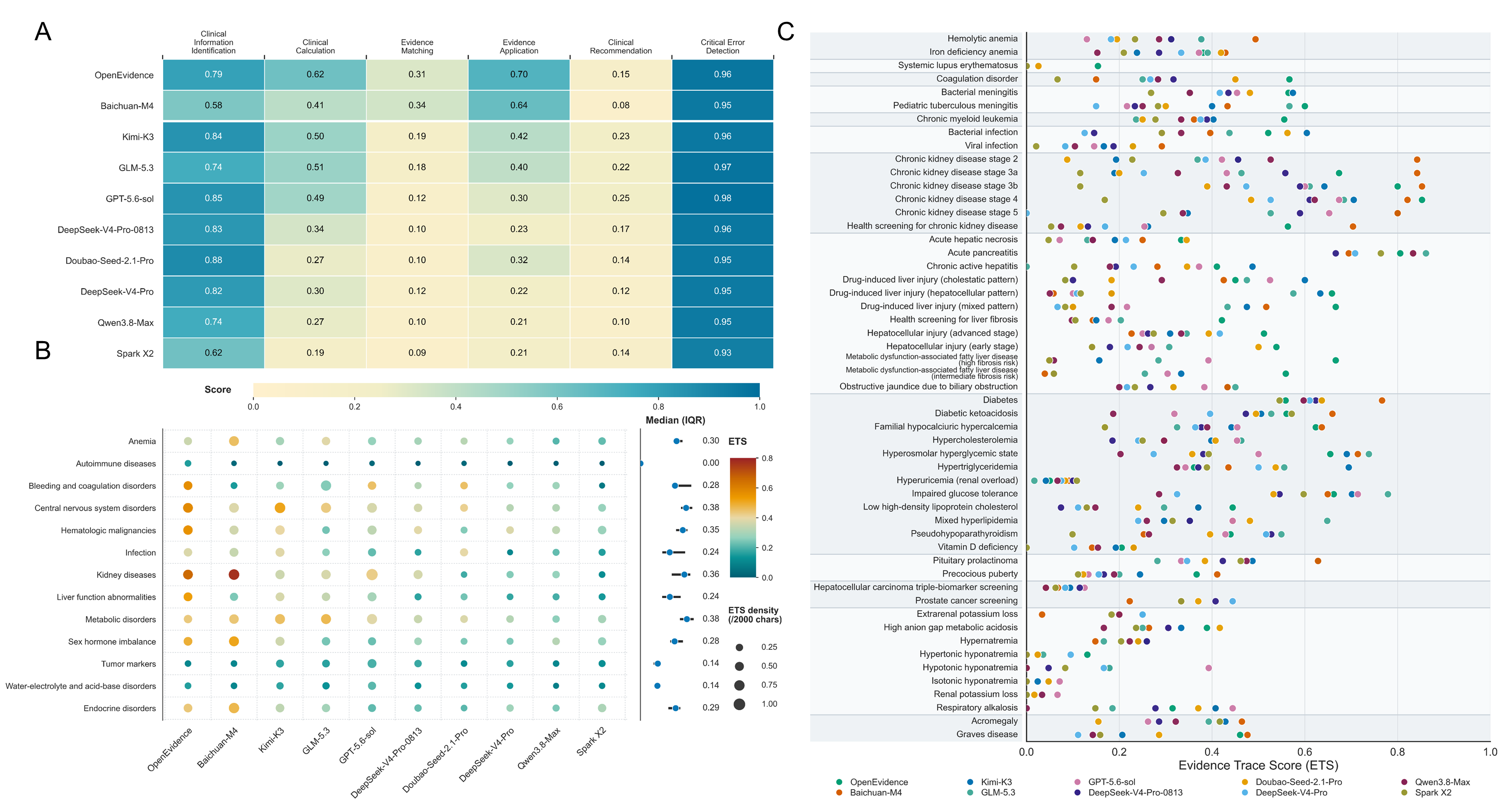}
\caption{\textbf{Performance on real-world clinical scenarios.} \textbf{a}, Scores of each model at the six CCEECC stages ($n = 365$ items; 2,992 criteria). The score for critical-error detection is one minus the proportion of available penalty points that were triggered. \textbf{b}, ETS by laboratory category. In the left panel, each marker represents one model in one category; color indicates the ETS and marker size indicates ETS density (ETS × 2,000 / chars). In the right panel, points and horizontal lines show the median and interquartile range of the ETS across the ten models. \textbf{c}, ETS for each of the 53 diseases or conditions. Each dot represents one model, and shaded bands group the scenarios by laboratory category.}
\label{fig:4}
\end{figure}}
\newcommand{\figV}[2]{%
\begin{figure}[#1]
\centering
\includegraphics[width=#2]{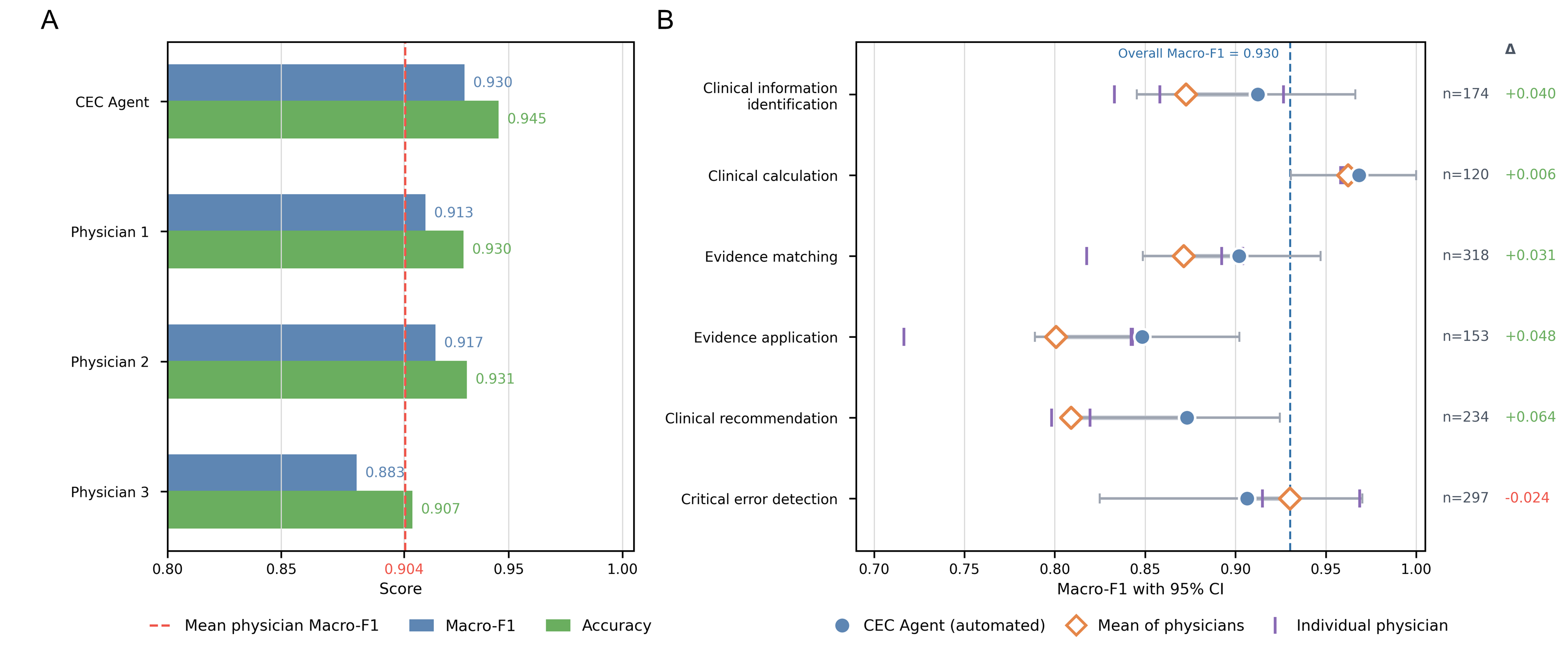}
\caption{\textbf{Agreement between automated and physician grading.} \textbf{a}, Macro-F1 (blue) and accuracy (green) of the CEC Agent against the majority vote of three physicians, and of each physician against a reference derived from the other two, with disagreements adjudicated by a fourth physician, on 159 real-world items (1,296 criterion-level judgments). The dashed red line shows the mean macro-F1 of the three physicians (0.904). \textbf{b}, Macro-F1 by CCEECC stage. Filled circles show the CEC Agent, open diamonds show the mean macro-F1 of the three physicians and vertical ticks show individual physicians; horizontal lines show 95\% CIs for the CEC Agent from 5,000 item-level cluster bootstrap resamples. The dashed blue line shows the overall macro-F1 of the CEC Agent (0.930). $n$, number of criterion-level judgments; $\Delta$, macro-F1 of the CEC Agent minus the mean macro-F1 of the physicians. Values are given in Extended Data Table~3.}
\label{fig:5}
\end{figure}}
\newcommand{\figVI}[2]{%
\begin{figure}[#1]
\centering
\includegraphics[width=#2]{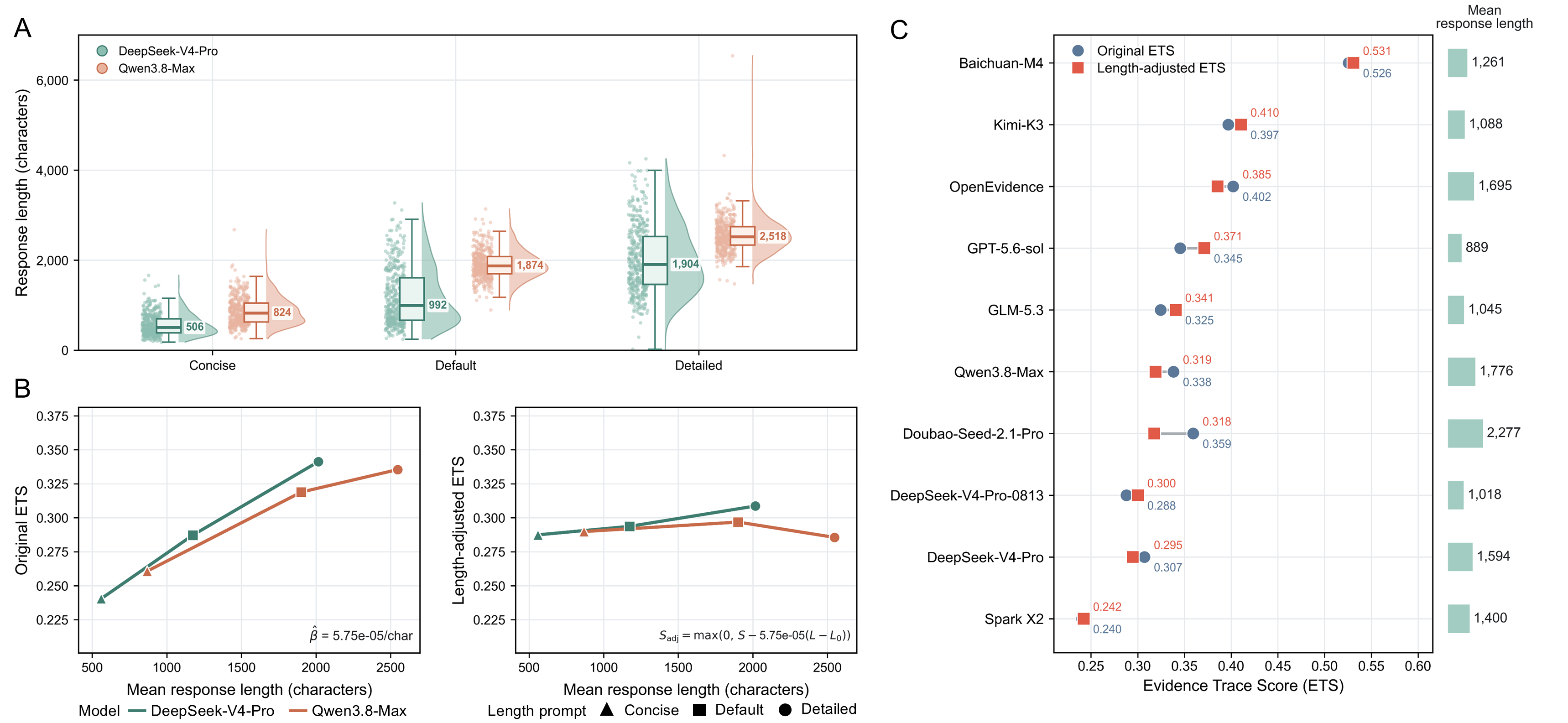}
\caption{\textbf{Sensitivity of the ETS to response length.} \textbf{a}, Response length of DeepSeek-V4-Pro and Qwen3.8-Max under prompts that requested a concise, default or detailed response. Each dot represents one response; boxes show the median and interquartile range, whiskers 1.5× that range and shaded curves the distribution. The median is given beside each box. \textbf{b}, Mean ETS against mean response length under the three prompts, before (left) and after (right) adjustment for length. Triangles, squares and circles denote the concise, default and detailed prompts, respectively.  \textbf{c}, ETS before (circles) and after (squares) adjustment for length for each of the ten models across the 1,217 open-ended tasks (standardized and real-world scenarios combined); bars at the right show the mean response length in characters. }
\label{fig:6}
\end{figure}}

\begin{document}

\title[Med Evidence Bench]{An evidence-traceable benchmark of clinical reasoning in large language models }

\author[1]{\fnm{Zhiqun} \sur{Zhao}}
\equalcont{These authors contributed equally to this work.}
\author[2]{\fnm{Xiaodong} \sur{Yang}}
\equalcont{These authors contributed equally to this work.}
\author[3,4]{\fnm{Bruce} \sur{Gu}}
\equalcont{These authors contributed equally to this work.}
\author[5]{\fnm{Ming} \sur{Li}}
\author[6]{\fnm{Yi} \sur{Fu}}
\author[7]{\fnm{Song} \sur{Li}}
\author[8]{\fnm{Qi} \sur{Mei}}
\author[9]{\fnm{Zhi} \sur{Liu}}
\author[10]{\fnm{Kai} \sur{Zhang}}
\author[3,4]{\fnm{Kexue} \sur{Fu}}
\author[11]{\fnm{Jing} \sur{Wu}}
\author[12]{\fnm{Xiaofei} \sur{Lu}}
\author[1]{\fnm{Yule} \sur{Li}}
\author*[2]{\fnm{Zheng} \sur{Wang}}\email{wangzheng@email.sdfmu.edu.cn}
\author*[1]{\fnm{Haiquan} \sur{Lu}}\email{lvhaiquan@sdu.edu.cn}
\author*[3,4]{\fnm{Xiaoming} \sur{Wu}}\email{wuxm@sdas.org}

\affil[1]{\orgdiv{Advanced Medical Research Institute and Key Laboratory for Experimental Teratology of the Ministry of Education, Cheeloo College of Medicine}, \orgname{Shandong University}, \orgaddress{\city{Jinan}, \country{China}}}
\affil[2]{\orgdiv{Department of Urology}, \orgname{Shandong Provincial Hospital Affiliated to Shandong First Medical University}, \orgaddress{\city{Jinan}, \country{China}}}
\affil[3]{\orgname{Shandong Computer Science Center (National Supercomputer Center in Jinan)}, \orgaddress{\city{Jinan}, \postcode{250014}, \country{China}}}
\affil[4]{\orgname{Qilu University of Technology (Shandong Academy of Sciences)}, \orgaddress{\city{Jinan}, \country{China}}}
\affil[5]{\orgdiv{Department of Rheumatology and Immunology}, \orgname{Shandong Provincial Hospital Affiliated to Shandong First Medical University}, \orgaddress{\city{Jinan}, \country{China}}}
\affil[6]{\orgdiv{Information Network Management Office}, \orgname{Shandong Provincial Hospital Affiliated to Shandong First Medical University}, \orgaddress{\city{Jinan}, \country{China}}}
\affil[7]{\orgdiv{Department of Medical Oncology}, \orgname{Qilu Hospital, Shandong University}, \orgaddress{\city{Jinan}, \country{China}}}
\affil[8]{\orgdiv{Department of Oncology}, \orgname{Tongji Hospital, Tongji Medical College, Huazhong University of Science and Technology}, \orgaddress{\city{Wuhan}, \country{China}}}
\affil[9]{\orgdiv{Department of Clinical Laboratory}, \orgname{Shandong Provincial Hospital Affiliated to Shandong First Medical University}, \orgaddress{\city{Jinan}, \country{China}}}
\affil[10]{\orgdiv{Department of Breast Surgery, General Surgery}, \orgname{Qilu Hospital, Shandong University}, \orgaddress{\city{Jinan}, \country{China}}}
\affil[11]{\orgdiv{Department of Central Laboratory}, \orgname{Public Health Clinical Center Affiliated to Shandong University, Shandong University}, \orgaddress{\city{Jinan}, \country{China}}}
\affil[12]{\orgdiv{Department of Breast Surgery}, \orgname{The First Affiliated Hospital of Shandong First Medical University}, \orgaddress{\city{Jinan}, \country{China}}}

\abstract{Benchmark scores for medical large language models (LLMs) are difficult to interpret when scoring criteria cannot be traced to clinical evidence. Here we developed Med Evidence Bench (MEB; \url{https://www.medevidencebench.org.cn/}), comprising 1,536 physician-constructed items across three levels: guideline knowledge, standardized clinical scenarios and real-world laboratory scenarios. Each criterion stating a medical claim is linked to a versioned source-text span. Across ten LLMs, the Evidence Trace Score (ETS) was 0.975--0.997 for guideline knowledge, 0.292--0.658 for standardized scenarios and 0.182--0.447 for real-world scenarios, and rankings varied across  levels. For real-world scenarios, scores for identifying clinical information (mean, 0.770) exceeded those for matching evidence (0.165) or formulating recommendations (0.160). On 1,296 criterion-level judgments, automated grading agreed with the majority vote of three physicians (macro-F1, 0.930; 95\% confidence interval, 0.915--0.945), although agreement was lower for evidence application. These findings from Chinese-language materials indicate that guideline recall does not ensure patient-specific evidence use and do not establish clinical effectiveness.}

\maketitle

\section*{Main}\label{sec:intro}

Large language models (LLMs) are increasingly evaluated for their ability to answer clinical questions, yet independent studies have reached opposing conclusions about which systems perform best. Vishwanath et al. reported that frontier general-purpose LLMs outperformed specialized clinical artificial intelligence (AI) tools on medical benchmarks\cite{vishwanath2026general}, a conclusion that has been questioned because of the limited scope of the benchmarks used\cite{beaulieujones2026limited}. Feng et al., using specialty-matched physician graders and real point-of-care queries, found that a specialized clinical tool scored highest\cite{feng2026expert}. More broadly, medical AI evaluation has been described as having a measurement problem\cite{manrai2026measurement}. The meaning of  a benchmark score depends on the construct being measured, the tasks used to elicit it and the way in which responses are scored\cite{bean2025measuring,bommasani2023holistic}. When these elements are incompletely specified, scores cannot be interpreted as evidence of clinical utility\cite{rajkomar2026validity}, which requires staged evaluation in clinical settings\cite{vasey2022decideai}.

The evaluation of medical LLMs has progressed from examination-style question answering\cite{jin2021medqa,singhal2023encode} to open-ended health conversations scored against physician-written rubrics\cite{arora2025healthbench}, frameworks that assess safety and effectiveness\cite{wang2026csedb}, stage-wise assessment of clinical reasoning\cite{qiu2025medrbench}, competency-based assessment in general practice\cite{li2026gpbench}, decision-making on real patient records\cite{hager2024evaluation} and evaluation suites spanning multiple medical tasks\cite{bedi2026medhelm}. Systematic reviews nonetheless describe substantial heterogeneity in how LLMs are evaluated\cite{shool2025systematic,laskar2024systematic}, a scarcity of evaluations based on data from real patient care\cite{bedi2025testing} and a gap between performance on knowledge-based examinations and on practice-oriented tasks\cite{gong2025gap}. We identify three remaining problems. First, the construct that a benchmark is intended to measure is often not defined explicitly. Second, the criteria used to score a response are seldom linked to the evidence that justifies them, preventing scores  being audited. Third, conclusions are sensitive to the choice of grader and protocol, which limits reproducibility.

Evidence-based medicine offers a way to address these problems by linking clinical judgments to explicit and verifiable evidence. Existing benchmarks have drawn on evidence-based medicine through tasks such as the extraction of trial elements and the summarization of randomized trials\cite{li2025ebm}, guideline-derived weighted criteria\cite{fast2024amega}, decision pathways derived from guidelines\cite{li2025medguide}, agreement with the conclusions of systematic reviews\cite{polzak2026medevidence} and question-answering tasks derived from systematic reviews and guidelines\cite{wang2026evaluating}. These benchmarks are organized around questions from the literature, predefined guideline nodes or reference conclusions. They do not test whether a model given individual patient data can identify the relevant evidence, establish its application to that patient, and calibrate the strength of its conclusion to the certainty of the evidence. We distinguish two requirements. Evidence application is the correct use of an evidence-derived rule when the clinical question is explicit and the necessary information is supplied. Evidence activation additionally requires the model to recognize an unstated clinical problem from patient data before applying evidence.

Here we present Med Evidence Bench (MEB), a benchmark that defines evidence-based clinical reasoning as its target construct and measures it at three levels: guideline knowledge; standardized clinical scenarios, which test evidence application; and real-world scenarios built from laboratory records, which test evidence activation (Fig.~\ref{fig:1}). Physicians decomposed each expected response into weighted, independently adjudicable criteria, and each criterion stating a medical claim is linked  to a versioned evidence record and the supporting span of source text. Responses were scored by an automated grader, the agreement of which with physicians was measured, and the sensitivity of scores to response length was examined. We used MEB to evaluate ten general-purpose and medical LLMs. We report variation in performance by evaluation level, clinical intent and capability domain, and we do not equate benchmark performance with clinical effectiveness or readiness for deployment.

\iffiguresinline\figI{!t}{\textwidth}\fi

\section*{Results}\label{sec:results}

\subsection*{Design and composition of MEB}\label{sec:r1}

MEB comprises 1,536 items at three evaluation levels (Fig.~\ref{fig:1}a). The 319 guideline-knowledge items are single-best-answer questions that test recall of a stated guideline recommendation. The 852 standardized clinical scenarios state the clinical question and provide the variables needed to answer it, thereby testing evidence application. The 365 real-world clinical scenarios each present a complete, de-identified laboratory report with limited clinical context; the clinical problem is not stated, allowing these items to test evidence activation. Items were classified by eight clinical intents, and criteria by four capability domains: clinical foundations, evidence-based decision-making, clinical reasoning and clinical safety (Extended Data Table~2).

Physicians decomposed the expected content of each response into atomic criteria that can be adjudicated independently, with weights that reflect clinical importance. The benchmark contains 6,411 criteria: 709 in clinical foundations, 2,950 in evidence-based decision-making, 1,185 in clinical reasoning and 1,567 in clinical safety. Each criterion stating a medical claim is linked to a versioned evidence record and the supporting span of source text, and the applicability of the evidence to the index case was verified in a physician-in-the-loop quality-control workflow (MEB-QC; Fig.~\ref{fig:1}b and Fig.~\ref{fig:2}). The capability domains were not fully crossed with the evaluation levels: each guideline-knowledge item had a single criterion in clinical foundations, and standardized scenarios had no criteria in clinical foundations. In real-world scenarios, the 2,992 criteria were additionally assigned to six stages of a reasoning chain termed CCEECC: clinical information identification (390 criteria), clinical calculation (281), evidence matching (734), evidence application (360), clinical recommendation (512) and critical-error detection (715).

\iffiguresinline\figII{!t}{0.74\textwidth}\fi

Diagnosis was the most common clinical intent (532 of 1,536 items, 34.6\%), followed by medication assessment (254, 16.5\%) and classification (244, 15.9\%); each of the remaining five intents accounted for 65--143 items (4.2--9.3\%). Guideline-knowledge items covered 15 specialties and standardized scenarios covered 12 clinical departments. Real-world scenarios covered 53 diseases or conditions in 13 laboratory  categories, the largest of which were metabolic disorders (90 items, 24.7\%), liver-function abnormalities (80, 21.9\%), water--electrolyte and acid--base disorders (54, 14.8\%) and kidney diseases (41, 11.2\%). Prompts for real-world scenarios were, on average, 7.1 times the length of those for standardized scenarios (mean, 666 versus 93 tokens).

\subsection*{Performance across evaluation levels}\label{sec:r2}

We evaluated ten LLM systems under fixed prompts: eight general-purpose models and two systems developed for medicine, Baichuan-M4 and OpenEvidence (Extended Data Table~1). The primary metric, the Evidence Trace Score (ETS), is the proportion of available rubric points earned by a model, with items in which any safety-critical error was triggered contributing no points (Methods). Across all 1,536 items, the ETS ranged from 0.302 for Spark X2 to 0.563 for Baichuan-M4; OpenEvidence (0.449) and Kimi-K3 (0.446) ranked second and third (Fig.~\ref{fig:3}a). Because the ETS is weighted by points, the overall score largely reflects performance on the two open-ended levels, which contributed 48.9\% and 42.8\% of the available positive points (standardized and real-world scenarios, respectively), compared with 8.3\% for guideline knowledge. 

\iffiguresinline\figIII{!t}{\textwidth}\fi

Scores differed markedly between levels (Fig.~\ref{fig:3}b). For guideline knowledge, the ETS ranged from 0.975 to 0.997, which corresponds to 311--318 of 319 items answered correctly. The ten models were thus separated by at most seven items, and their 95\% confidence intervals (CIs) overlapped (0.951--0.987 for the lowest-scoring model and 0.982--0.999 for the highest-scoring model). The ETS ranged from 0.292 to 0.658 for standardized scenarios and from 0.182 to 0.447 for real-world scenarios. Eight models scored lower on real-world scenarios than on standardized scenarios, whereas OpenEvidence scored higher (0.447 versus 0.363) and GLM-5.3 had similar scores on both (0.328 and 0.322). Because the levels differ in response format, in the composition of the rubric and in the information supplied, these differences cannot be attributed to task difficulty alone.

The ranking of models differed between the two open-ended levels (Spearman's $\rho = 0.19$; 95\% CI, $-0.51$ to 0.73; $n = 10$ models); the wide interval indicates that concordance between levels cannot be estimated precisely from ten models. Baichuan-M4 ranked first on standardized scenarios (0.658) and second on real-world scenarios (0.374), whereas OpenEvidence ranked seventh (0.363) and first (0.447), respectively. Among the general-purpose models, Doubao-Seed-2.1-Pro and Qwen3.8-Max ranked second and third on standardized scenarios (0.453 and 0.448) but seventh and ninth on real-world scenarios (0.252 and 0.213), and Kimi-K3 ranked fourth and third (0.442 and 0.345). Rank correlations that involved guideline knowledge ($\rho = -0.18$ with standardized scenarios and $-0.40$ with real-world scenarios) are not informative, because scores at that level were near ceiling. The two medical systems ranked first and second overall. Baichuan-M4 led standardized scenarios, whereas OpenEvidence led real-world scenarios. However, only two such systems were evaluated, and OpenEvidence was accessed through its web interface rather than under the conditions applied to the other models; these comparisons therefore do not support conclusions about medical systems as a class.

\subsection*{Capability domains and reasoning stages}\label{sec:r3}

Across all items, the mean score for the ten models was 0.976 (range, 0.962--0.988) for clinical safety, 0.897 (0.815--0.943) for clinical foundations, 0.543 (0.384--0.718) for clinical reasoning and 0.236 (0.153--0.459) for evidence-based decision-making (Fig.~\ref{fig:3}c). OpenEvidence had the highest score for clinical reasoning (0.718) and Baichuan-M4 had the highest score for evidence-based decision-making (0.459). These scores should be interpreted in light of the criteria they include: the clinical-foundations domain pools guideline-knowledge items, on which all models scored near ceiling, with criteria for the identification of clinical information in real-world scenarios.

Decomposition of real-world scenarios into six stages identified where models lost points (Fig.~\ref{fig:4}a). Mean scores were 0.770 (range, 0.583--0.880) for clinical information identification, 0.392 (0.191--0.624) for clinical calculation, 0.165 (0.090--0.344) for evidence matching, 0.366 (0.207--0.702) for evidence application and 0.160 (0.082--0.246) for clinical recommendation. The score for critical-error detection, which reflects the absence of enumerated safety-critical errors, was 0.955 (0.926--0.982). Scores did not decline monotonically along the chain: in every model, the score for evidence application exceeded that for evidence matching. No model led at every stage. OpenEvidence had the highest scores for clinical calculation (0.624) and evidence application (0.702); Baichuan-M4 had the highest score for evidence matching (0.344) but the lowest scores for clinical information identification (0.583) and clinical recommendation (0.082); and GPT-5.6-sol had the highest score for clinical recommendation (0.246). Criteria for evidence matching required responses to state the applicable evidence-derived rule and name a supporting source.

\iffiguresinline\figIV{!t}{\textwidth}\fi

\subsection*{Variation across clinical categories and intents}\label{sec:r4}

Results stratified by clinical intent across all items are shown in Extended Data Fig.~1. Among standardized scenarios, obstetrics and gynecology was the lowest-scoring department in nine of the ten models (0.124--0.262); the exception was Baichuan-M4, which scored 0.533 in obstetrics and gynecology and lowest in neurology (0.483) (Extended Data Fig.~2). When real-world scenarios were grouped by clinical intent, scores were low in every model for classification (0.118--0.236) and cancer screening (0.102--0.195; the same 18 tumor-marker items), and OpenEvidence had the highest ETS for risk stratification (0.657), medication assessment (0.592) and emergency triage (0.579).

Performance on real-world scenarios varied across the 13 laboratory categories (Fig.~\ref{fig:4}b). The mean ETS across models was lowest for autoimmune diseases (0.023; range, 0--0.154; $n = 6$ items), water--electrolyte and acid--base disorders (0.140; 0.083--0.210; $n = 54$) and tumor markers (0.142; 0.102--0.195; $n = 18$). Differences between models were largest for kidney diseases ($n = 41$), for which the ETS ranged from 0.121 to 0.772 and was highest for Baichuan-M4 (0.772) and OpenEvidence (0.658). The highest-scoring model varied by category: OpenEvidence in five categories, Baichuan-M4 in four, GPT-5.6-sol in two, and GLM-5.3 and Doubao-Seed-2.1-Pro in one each. Among the 53 diseases or conditions (Fig.~\ref{fig:4}c), the ETS for acute pancreatitis was 0.667--0.861 across models, whereas that for drug-induced liver injury of hepatocellular pattern ranged from 0.050 to 0.658. Each disease or condition comprised few items, and results at this level are therefore descriptive.

\subsection*{Agreement between automated and physician grading}\label{sec:r5}

Automated grading by the CEC Agent was compared with grading by three physician reviewers who had not participated in developing the rubrics. The validation set comprised 159 items from the real-world subset, with three items sampled at random from each of the 53 diseases or conditions; the responses of DeepSeek-V4-Pro to these items comprised 1,296 criterion-level judgments. Against the majority vote of the three physicians, the CEC Agent had a macro-F1 of 0.930 (95\% CI, 0.915--0.945), an accuracy of 0.945 and a Cohen's $\kappa$ of 0.861. For comparison, each physician was assessed against a reference derived from the other two physicians, with a fourth physician adjudicating criteria on which they disagreed; the mean macro-F1 was 0.904 (95\% CI, 0.888--0.919), the mean accuracy was 0.923 and the mean $\kappa$ was 0.810 (Fig.~\ref{fig:5}a and Extended Data Table~3).

\iffiguresinline\figV{!t}{\textwidth}\fi

Agreement differed by stage (Fig.~\ref{fig:5}b). The macro-F1 of the CEC Agent was 0.968 (95\% CI, 0.931--1.000) for clinical calculation, 0.912 (0.845--0.966) for clinical information identification, 0.906 (0.825--0.970) for critical-error detection, 0.873 (0.811--0.925) for clinical recommendation, 0.902 (0.849--0.947) for evidence matching and 0.848 (0.789--0.902) for evidence application, for which Cohen's $\kappa$ was 0.697. The corresponding means for physicians were 0.962, 0.873, 0.930, 0.809, 0.871 and 0.801. Agreement among physicians was lowest for evidence application: the Cohen's $\kappa$ of individual physicians was 0.64--0.81 for evidence matching and 0.48--0.69 for evidence application. Few criteria were judged to be met at the evidence-matching stage, so the stage-specific estimates are based on few positive labels and are imprecise. The higher values of the CEC Agent at the two evidence stages should therefore be interpreted in the context of ambiguous adjudication and do not indicate an advantage of automated grading. The automated grader was not validated on standardized scenarios.

\subsection*{Sensitivity of scores to response length}\label{sec:r6}

Because a longer response has more opportunities to satisfy rubric criteria, we examined the relationship between scores and response length by prompting two models, DeepSeek-V4-Pro and Qwen3.8-Max, to provide concise, default or detailed responses on 284 standardized and 122 real-world clinical scenarios. Across these items, median response length increased from 505.5 to 1,904.5 characters for DeepSeek-V4-Pro and from 823.5 to 2,518.5 characters for Qwen3.8-Max, and the ETS increased with response length in both models (Fig.~\ref{fig:6}a,b). Averaging the four slopes with equal weight on the two levels gave a coefficient of $5.75 \times 10^{-5}$ per character, which corresponds to an increase in the ETS of 0.058 per 1,000 characters. Because verbosity was manipulated through the prompt, this slope reflects the combined effect of the instruction on the length and the content of responses.

\iffiguresinline\figVI{!t}{\textwidth}\fi

We used this coefficient, with a reference length specific to each level, to calculate a length-adjusted ETS for each of the ten models on all 852 standardized and 365 real-world clinical scenarios (Fig.~\ref{fig:6}c and Extended Data Table~4). The ranking was unchanged for real-world scenarios (Spearman's $\rho = 1.00$) and largely preserved for standardized scenarios (Spearman's $\rho = 0.93$). For real-world scenarios, OpenEvidence remained first (0.447 before and 0.424 after adjustment), Baichuan-M4 second (0.374 and 0.386) and Kimi-K3 third (0.345 and 0.358). For standardized scenarios, Baichuan-M4 remained first (0.658 before and after adjustment) and Qwen3.8-Max remained third (0.448 and 0.426); second place changed from Doubao-Seed-2.1-Pro (0.453 before and 0.407 after adjustment) to Kimi-K3 (0.442 and 0.456), and fifth place changed from DeepSeek-V4-Pro to GPT-5.6-sol (Extended Data Table~4). Because the slope was estimated in two models and applied to all ten, the adjusted scores constitute a sensitivity analysis and do not remove the influence of response length.

\FloatBarrier

\section*{Discussion}\label{sec:discussion}

We developed a benchmark in which every scoring criterion can be traced to a span of source text, and used it to evaluate ten LLMs at three levels that differ in the information provided to the model. All ten models answered almost all guideline-knowledge questions correctly, but earned fewer than two thirds of the available points when required to apply evidence to explicit clinical scenarios, and fewer than half when required to work from complete laboratory reports. In real-world scenarios, the models usually identified the relevant abnormalities but seldom stated the applicable evidence or formulated the recommendations specified by physicians. Model rankings differed between levels, and no model led at every stage of reasoning.

These findings are consistent with the gap between knowledge-based and practice-based benchmarks that was described in a systematic review of 39 benchmarks\cite{gong2025gap}, with the lower performance reported for the recommendation of examinations and the planning of treatment than for diagnosis\cite{qiu2025medrbench}, and with the difficulties that LLMs have shown in following guidelines and interpreting laboratory results when working from real patient records\cite{hager2024evaluation}. Evaluation in simulated patient interactions has likewise shown that performance on examination-style vignettes does not carry over to more realistic formats\cite{johri2025craftmd}. Differences between levels in the present study should nevertheless be interpreted cautiously. The levels differ in the information provided, response format and rubric composition, and guideline-knowledge items named the guideline to be recalled. A score near ceiling at that level shows that the content of the guidelines was accessible to the models.

The low scores for evidence matching and for evidence-based decision-making require careful interpretation. Criteria at the evidence-matching stage required responses to state the evidence-derived rule and name a supporting source, most often a Chinese guideline or consensus statement, whereas the prompts asked for an interpretation of the case. A response could therefore reach a correct, case-specific conclusion and still lose the points assigned to explicit attribution, which is consistent with the higher scores for evidence application than for evidence matching in every model. We regard explicit attribution as a legitimate component of evidence-based reasoning, because it allows a clinician to verify the basis of a recommendation. However, scores at this stage reflect the tendency of a model to attribute its statements without being prompted, as well as its ability to select the appropriate evidence.

Our results inform the comparison between general-purpose and specialized medical systems but do not resolve it. The two medical systems had the highest overall scores and led both open-ended levels, a pattern that has also been reported in evaluations based on clinical criteria\cite{feng2026expert,wang2026csedb} and that differs from the pattern reported on medical benchmarks\cite{vishwanath2026general}. However, only two medical systems were evaluated, their advantage was not uniform across levels, stages or clinical categories, and one of them was accessed under conditions that differed from those of the other models. More generally, the dependence of rankings on the evaluation level indicates that the information supplied to a model is part of what is being estimated, and that a single aggregate score can conceal differences relevant to a particular use. Benchmarks that span several tasks have similarly shown that the performance of LLMs varies by task and setting\cite{wu2025versatile,chen2025benchmarking,bianchi2026cardbiomedbench}.

A second contribution of MEB is that the basis of each score can be inspected. Each criterion stating a medical claim is linked to a versioned chain from criterion to evidence record and source-text span, and the applicability of the evidence to the index case was verified rather than inferred from the citation of a guideline. This design follows principles of dataset documentation\cite{gebru2021datasheets} and complements frameworks for the human evaluation of LLMs in healthcare\cite{tam2024framework}. It allows the criteria affected by the revision of a guideline to be identified and re-audited, although we have not yet evaluated this process across successive versions. MEB also scores safety separately from overall correctness, through enumerated errors such as the alteration of source data, the misuse of thresholds and the treatment of untested values as valid inputs, an approach related to frameworks that classify errors in LLM-generated clinical documentation\cite{asgari2025framework,seo2024framework} and to benchmarks of the detection of medical errors in clinical notes\cite{benabacha2025medec}. The high safety scores observed here indicate that these specific errors were uncommon; they do not indicate that the responses would be safe in clinical use. The low scores for evidence matching and clinical recommendation suggest that methods intended to ground responses in evidence, such as retrieval-augmented generation\cite{xiong2024benchmarking,amugongo2025rag,wang2025accelerating}, should be assessed not only for retrieval but also for whether the retrieved evidence applies to the case and is translated into an appropriate recommendation; we did not evaluate retrieval-augmented configurations of the general-purpose models.

Automated grading made evaluation on this scale feasible, and its overall agreement with the majority vote of three physicians was similar to that of individual physicians. Three caveats apply. First, agreement among physicians was lowest for evidence application, and the evidence-matching estimates were based on few positive labels; scores at these stages therefore carry greater measurement uncertainty. Second, validation was confined to real-world scenarios. Third, agreement with physicians does not exclude biases that an LLM-based grader may share with the models that it grades\cite{zheng2023judging,panickssery2024llm}. Our findings align with studies in which LLM-based graders approximated expert ratings of clinical text\cite{croxford2025judges,zhou2026automating}, and they support the use of automated grading as an approximation under the tested rubric  rather than as a substitute for adjudication by physicians. Scores also increased with response length when verbosity was manipulated through the prompt. Adjustment for length left the ranking unchanged for real-world scenarios and largely preserved for standardized scenarios, but the adjustment relies on a slope estimated in two models, following approaches developed for other benchmarks\cite{hicks2026healthbenchpro,dubois2024length}.

This study has several limitations. First, the real-world scenarios were built from the records of patients in China, the evidence corpus consisted predominantly of Chinese guidelines and consensus statements, and all materials were administered and scored in Chinese. Scores may therefore reflect familiarity with Chinese terminology, reference intervals, care pathways and guideline conventions, and responses that relied on equivalent international guidance may not have received credit; the findings may not generalize to other languages, health systems, populations or assay platforms. Second, the models were not compared under identical conditions: most were accessed through application programming interfaces with the default sampling parameters of each provider, whereas OpenEvidence was accessed through its web interface, and each item was submitted once, so that variability between runs was not estimated. Third, the levels were not matched on clinical content and the capability domains were not fully crossed with the levels, so that differences between levels and between domains are partly attributable to the composition of the rubric. Fourth, estimates for small subgroups, including individual laboratory scenarios, are imprecise, and differences between models were not formally tested. Fifth, the performance of physicians on MEB was not measured, so that the scores of the models cannot be compared with those of clinicians who answer under the same conditions. Sixth, MEB assesses the interpretation of laboratory data in a single turn and does not cover imaging, longitudinal trajectories, multi-turn dialogue or the use of tools by agent-based systems\cite{jiang2025medagentbench,liu2026agents}; nor does it assess performance across patient subgroups\cite{templin2025bias}. Finally, benchmark evaluation cannot establish safety, effectiveness or benefit for patients, which require prospective evaluation in clinical settings\cite{vasey2022decideai,schaekermann2026prospective}.

In summary, MEB distinguishes guideline recall, evidence application to explicit clinical questions and evidence activation from patient data, and makes the basis of each score inspectable. Under the conditions tested, near-perfect recall of guidelines was accompanied by limited use of evidence for individual patients. Reporting performance separately by information condition, reasoning stage and clinical category, together with the evidence behind each criterion, may make evaluations of medical LLMs easier to interpret and compare, in line with calls for more rigorous tests of medical AI\cite{goh2026superintelligence}.

\section*{Methods}\label{sec:methods}

\subsection*{Study design}\label{sec:m1}
We developed MEB and used it in a cross-sectional evaluation of ten LLM systems. The study comprised five components: construction of the benchmark by physicians, with quality control supported by an LLM-based agent (MEB-QC); generation of responses by the evaluated models; automated criterion-level grading (CEC Agent); validation of the automated grader against physician reviewers; and analysis of the sensitivity of scores to response length. The components of the evaluation, including the items, criteria and weights, the hierarchy of evidence sources, the safety rules, the prompts, the grading procedure, the aggregation of scores and the adjustment for response length, were fixed on 1 May 2026  before the models were evaluated and were applied identically to all models. Reporting follows the TRIPOD-LLM statement\cite{gallifant2025tripodllm} and the MI-CLAIM-GEN checklist\cite{miao2025miclaimgen}.

\subsection*{Evaluation levels}\label{sec:m3}
MEB has three levels, which progress from single-best-answer questions to the open-ended interpretation of clinical data and place increasing demands on the application of evidence and the integration of information. All items were administered and scored in Chinese.

\textbf{Guideline knowledge.} Guideline knowledge was assessed with single-best-answer multiple-choice questions derived from current authoritative guidelines. Each item addressed one clinical decision and comprised a stem, one keyed answer with a citation of the source guideline, and distractors that represented common clinical misconceptions, outdated recommendations or advice that does not apply to the specified population. Clinical experts wrote the items and verified the keyed answers against the source recommendations. The 319 items covered 15 specialties.

\textbf{Standardized clinical scenarios.} To measure evidence application when the information is sufficient, the clinical question is explicit and distracting information is minimal, physicians reviewed current guidelines and expert consensus statements and identified evidence-derived rules that had a defined target population, input variables, decision thresholds and recommended actions. Each rule was used to construct a scenario that contained only the information required for the target judgment, presented in a single prompt in natural language. The 852 scenarios covered 12 clinical departments, including 432 rare disease-related criteria.

\textbf{Real-world clinical scenarios.} Standardized scenarios do not reproduce the redundant information, co-occurring abnormalities and incomplete background that are typical of clinical records. To assess evidence activation, we therefore constructed scenarios from laboratory records. Each item retained the structured test items, results, units and reference intervals of the original report, including items that were not relevant to the target judgment, together with background information on the patient that was relevant to the judgment, when available. The 365 items covered 53 diseases or conditions  in 13 laboratory categories.

\subsection*{Clinical intents and capability domains}\label{sec:m4}
Items were classified by eight clinical intents: pre-consultation, diagnosis, classification, risk stratification, emergency triage, medication assessment, chronic disease management and cancer screening (Extended Data Table~2). 

Each criterion was assigned to one of four capability domains. Clinical foundations comprised knowledge of current guidelines, consensus statements and normative medical knowledge, and the identification of decision-relevant clinical information and abnormal findings in case materials. Evidence-based decision-making comprised the identification, matching and tracing of authoritative evidence, including diagnostic criteria, rules for classification or staging, clinical thresholds and the basis of recommendations, together with judgment of the applicability of the evidence to the index patient and, in real-world scenarios, the formulation of next-step recommendations on the basis of that evidence. Clinical reasoning comprised the integration of clinical information with applicable evidence, the calculation of composite indices, corrected indices or clinical scores, and case-specific judgments concerning diagnosis, differential diagnosis, classification, staging or risk. Clinical safety comprised the avoidance of safety-critical errors, including the misuse of thresholds or classification rules, the omission of required calculations or corrections, the alteration of source data, the treatment of missing or untested data as valid inputs, the presentation of conditional evidence as a definitive diagnosis, and recommendations that conflict with key evidence or could cause harm.

\subsection*{CCEECC reasoning stages}\label{sec:m5}
For real-world scenarios, the target reasoning process was decomposed into six stages that can be assessed independently.
(1) Clinical information identification: identification of the clinical information and abnormal results that are relevant to the clinical problem.
(2) Clinical calculation: recognition that a composite index, corrected index or clinical score can be derived from the available information, and correct calculation and interpretation against thresholds.
(3) Evidence matching: integration of the reported results and calculated values to recognize a clinically meaningful pattern of abnormalities, disease phenotype or risk state, and matching of that pattern to guidelines, consensus statements or other authoritative evidence that can adjudicate the question.
(4) Evidence application: judgment of the applicability of candidate evidence to the case, and application of the corresponding rule to reach a judgment on classification, staging, risk or diagnosis.
(5) Clinical recommendation: proposal of next steps that are proportionate to that judgment, including further or repeat testing, differential work-up, referral, treatment, management of medication or follow-up.
(6) Critical-error detection: a check of the response for enumerated safety-critical errors that could invalidate the judgment or increase risk to the patient, including the alteration of reported values, the misuse of key thresholds or classification rules, the omission of required calculations or corrections, the treatment of missing or untested values as valid inputs, and conclusions that conflict with key evidence. This stage is scored by the grader and records the absence of such errors in the response; it does not require the model to detect errors in the input.

In the counts of criteria, criteria for clinical information identification were assigned to clinical foundations; those for clinical calculation and evidence application to clinical reasoning; those for evidence matching and clinical recommendation to evidence-based decision-making; and those for critical-error detection to clinical safety.

\subsection*{Development of rubrics by physicians}\label{sec:m6}
Fourteen clinicians from seven tertiary hospitals in China, representing ten specialties, contributed in four roles: medical advisors ($n = 2$), who defined the scope of the tasks and the principles of annotation; specialist annotators ($n = 7$), who wrote the item-specific criteria; evidence verifiers ($n = 3$), who used the MEB-QC workflow to audit the evidence and its applicability; and independent reviewers ($n = 2$), who adjudicated disputed criteria. The purpose of annotation was to create units of measurement that can be adjudicated independently, and not only to produce reference answers. The following principles were applied.

First, each criterion described one expected or prohibited behavior. Composite judgments were decomposed into the smallest units that could be scored, and the capability domain, the conditions of applicability and the provenance of the evidence were specified for each item. Second, for real-world scenarios, physicians first annotated the primary and associated abnormalities and then identified clinically relevant patterns across analytes and composite indices that required calculation. Third, each criterion stating a medical claim  was linked to at least one traceable source of evidence. Weights reflected clinical importance and the potential harm of an error, and were independent of the expected length or wording of a response. Fourth, the wording of criteria was calibrated to the certainty of the evidence, and distinguished expressions such as `suggestive of', `should be considered', `highly suspicious for' and `diagnostic of'. Fifth, a medical claim was not scored more than once within an item; criteria with causal or inclusion relationships were merged or further decomposed so that each could be adjudicated independently. Sixth, all physicians involved in annotation, review and adjudication declared potential conflicts of interest concerning the evaluated models, related products and guideline-development organizations, and a physician with a direct conflict did not take part in the final adjudication of the corresponding criterion.

\subsection*{Evidence corpus and traceability}\label{sec:m7}
Criteria were supported by a controlled corpus of approximately 40,000 documents, which comprised guidelines, consensus statements, specifications, standards and authoritative textbooks\AQ{27}. Content from the open web that had not been vetted was not accepted as evidence. Sources were organized in three tiers defined by the study team: tier A comprised current international or national evidence-based guidelines and regulatory safety documents; tier B comprised peer-reviewed professional consensus statements; and tier C comprised current authoritative textbooks and institutionally validated standard operating procedures. This tiering was informed by established approaches to rating the certainty of evidence\cite{guyatt2025coregrade} and to appraising the quality of guidelines\cite{brouwers2010agree}, but is not itself a validated grading system. The tier of the source, the certainty of evidence as reported by the source and the strength of the recommendation were stored as separate fields and were not inferred automatically.

Each criterion stating a medical claim was linked to a versioned evidence record that contained, as applicable, the issuing organization, the title, year and version of the document, the span of source text, the target population, the clinical state, the test or variable, the specimen, the assay, the unit, the reference interval, the threshold or condition, the recommendation, the region, the certainty of evidence. The resulting chain from criterion to medical claim, evidence and source-text span makes the basis of each score auditable\cite{gebru2021datasheets}. Versioned provenance also allows the criteria affected by a change in a guideline to be identified and re-audited.

\subsection*{MEB-QC workflow}\label{sec:m8}
MEB-QC is a quality-control procedure that operates within the controlled evidence corpus. It supported physicians by detecting possible problems in the provenance of evidence, contextual applicability, calculations, units, thresholds and safety; it did not replace the judgment of physicians, establish clinical ground truth or make decisions on inclusion. Physicians submitted candidate cases, criteria and initial supporting evidence for audit. After the MEB-QC Agent had returned a report, physicians reviewed each finding and resubmitted revised material, and the cycle was repeated until every included criterion had a complete and applicable evidence chain, the numerical values, units and thresholds had been verified, and physicians had confirmed the accuracy of the case, the criteria and the supporting evidence. Independent experts adjudicated conflicts between guidelines and ambiguous evidence.

\textbf{Contextual applicability and inclusion of criteria.} Contextual applicability was defined as consistency between the clinical case and the conditions under which the evidence supporting a criterion applies. The citation of a guideline was not by itself accepted as showing that its recommendation applied to the case. The review of applicability considered the target population and age range; the clinical state, symptoms, comorbidities, medical history and medication history; the intervention or test; the type of specimen; the assay; the unit of measurement; the reference interval; the threshold; and the country, region and version of the document. A criterion was included only when its medical claim could be linked to a versioned evidence record and a source location that could be inspected. Tier A sources were given priority. The latest valid version of a document was used unless use of an earlier version was justified and it was confirmed that the differences between versions did not alter the criterion. Criteria preserved the distinctions made in the source document regarding the certainty of evidence and the strength of recommendations. When several current guidelines or consensus statements addressed the same question, the relevant sources in the corpus, including those from different regions, were compared with respect to their populations, conditions of applicability and recommendations, and independent experts adjudicated ambiguity that remained unresolved. For criteria that involved patterns across analytes, composite indices, conversions of units, clinical scores or changes over time, the audit checked the formulas, units, direction of thresholds, reference intervals and compatibility of assays against the supporting evidence.

\textbf{Architecture of the MEB-QC Agent.} The MEB-QC Agent comprised seven modules connected through a shared state graph (Fig.~\ref{fig:2}a). The Task Complexity Router assigned the depth of the audit according to the number of claims, the type of evidence, the need for comparison between guidelines, the presence of numerical calculations or safety-critical content, and the involvement of rare diseases. The Calculator performed arithmetic operations, conversions of units, calculations of clinical scores and comparisons with thresholds. The Claim--Evidence Graph Builder decomposed the case, the criteria and the accompanying explanatory text into atomic medical claims and generated structured queries conditioned on the clinical scenario. The Hybrid Retriever combined lexical and semantic retrieval with filtering on metadata and reranking with a cross-encoder, and referred uncertain conclusions to physicians. The Applicability Verifier compared the case, criteria, claims and retrieved evidence item by item for conflicts. The End-to-End Traceability Module linked criterion to a medical claim, an evidence record and a span of source text. The Audit Reporter returned a structured report in which each criterion was classified as verified, applicability mismatch or insufficient evidence, with a rationale that cited the evidence. The language-model calls in the MEB-QC Agent used gpt-oss-120b, the same model as the CEC Agent and not one of the evaluated systems. For those calls the temperature was set to 0, top-$p$ to 1.0, the reasoning effort to medium and the maximum output length to 4,096 tokens. 

\subsection*{Models and generation of responses}\label{sec:m9}
Ten LLM systems were evaluated (Extended Data Table~1): Baichuan-M4, DeepSeek-V4-Pro (preview version), DeepSeek-V4-Pro-0813, Doubao-Seed-2.1-Pro, GLM-5.3, GPT-5.6-sol, Kimi-K3, OpenEvidence, Qwen3.8-Max and Spark X2. Baichuan-M4 and OpenEvidence were classified as medical systems and the others as general-purpose models. Responses were generated with fixed user prompts. The fixed user prompts will be released with the peer-reviewed article. The maximum length of the output was set to 10,000 tokens; other sampling parameters were left at the default values of each provider, and reasoning modes were used as listed in Extended Data Table~1. 

\subsection*{Automated grading with the CEC Agent}\label{sec:m10}
Open-ended responses were scored with the Clinical calculation--Evidence--Clinical recommendation (CEC) Agent, a rubric-based LLM-as-Judge\cite{zheng2023judging}. The judge was gpt-oss-120b. It received the item prompt, the model response and the item-specific rubric, and classified each criterion independently as met or not met. The temperature was set to 1.0, top-$p$ to 1.0 and the reasoning effort to medium, and the maximum output length to 4,096 tokens. The grading prompts will be released with the peer-reviewed article. The CEC framework was a scaffold for scoring and was not interpreted as evidence of the internal reasoning of the evaluated model. Whereas the CCEECC chain specifies the target stages of reasoning, the scaffold parses a response along three aspects: clinical calculation (whether source data, units, reference intervals and temporal relationships were preserved and, when required, whether the formula, unit conversion, numerical result and threshold interpretation were correct); evidence (whether claims were matched to current rules whose conditions of applicability fit the case, rather than by citation of a guideline name alone, and whether the strength of the conclusion was calibrated to the certainty of the evidence); and clinical recommendation (whether proposed actions were commensurate with the inferred risk and the available evidence).

The CEC Agent comprised four modules. The Task Router determined from the criteria which aspects were to be assessed; the Response Parser represented the response in structured form; the Adjudicator returned a judgment and a rationale for each criterion, independently of the other criteria; and the Reporter compiled an auditable record of each criterion, its judgment and its rationale. Logs of each run recorded the names and versions of the evaluated model and the judge, the date, the prompts, the sampling parameters and the checkpoints used to resume interrupted runs.

\subsection*{Evidence Trace Score and domain scores}\label{sec:m11}
For item $i$, let $R_i^{+}$ denote the set of criteria that describe expected behaviors and $R_i^{-}$ the set of safety-critical criteria that describe prohibited behaviors, and let $w_{ir} > 0$ denote the weight of criterion $r$, which for a safety-critical criterion is the magnitude of the penalty. For model $m$, let $\hat{y}_{mir} \in \{0,1\}$ denote the judgment of the grader, where $\hat{y}_{mir} = 1$ indicates that an expected behavior was present or that a safety-critical error was triggered. The points earned, the penalty incurred and the maximum points available were defined as
\begin{equation}
P_{mi}=\sum_{r\in R_i^{+}} w_{ir}\,\hat{y}_{mir},\qquad
E_{mi}=\sum_{r\in R_i^{-}} w_{ir}\,\hat{y}_{mir},\qquad
D_{i}=\sum_{r\in R_i^{+}} w_{ir}.
\label{eq:points}
\end{equation}
The ETS of model $m$ over a set of items $G$ was the ratio of the points earned to the maximum points available, with an item in which any safety-critical error was triggered contributing no points to the numerator:
\begin{equation}
\mathrm{ETS}_{m}=\frac{\sum_{i\in G}\mathbf{1}[E_{mi}=0]\,P_{mi}}{\sum_{i\in G} D_{i}}.
\label{eq:ets}
\end{equation}
The ETS lies between 0 and 1 and was the primary metric. It was calculated over all items and separately for each evaluation level, clinical intent, department, laboratory category and laboratory scenario. For guideline-knowledge items, each of which carried a single criterion of equal weight, the ETS equals the proportion of items answered correctly.

For the capability domains of clinical foundations, evidence-based decision-making and clinical reasoning, and for the first five CCEECC stages, the score of model $m$ in domain or stage $d$ was computed from the criteria assigned to $d$, denoted $R_{i,d}^{+}$:
\begin{equation}
S_{md}=\frac{\sum_{i\in G}\sum_{r\in R_{i,d}^{+}} w_{ir}\,\hat{y}_{mir}}{\sum_{i\in G}\sum_{r\in R_{i,d}^{+}} w_{ir}}.
\label{eq:domain}
\end{equation}
The score for clinical safety, and that for critical-error detection in real-world scenarios, was one minus the proportion of available penalty points that were triggered:
\begin{equation}
S_{m}^{\mathrm{safety}}=1-\frac{\sum_{i\in G} E_{mi}}{\sum_{i\in G}\sum_{r\in R_i^{-}} w_{ir}}.
\label{eq:safety}
\end{equation}
This score summarizes the enumerated safety-critical criteria that were triggered under the present protocol and should not be interpreted as evidence of safety in clinical use.

\subsection*{Validation of automated grading}\label{sec:m12}
The validation set comprised three items sampled at random from each of the 53 diseases or conditions in the real-world subset ($n = 159$ items). For each item, the response generated by DeepSeek-V4-Pro was independently assessed by three senior physicians holding the rank of chief physician in China, none of whom had participated in developing the rubrics. The reviewers judged whether each criterion was met and were blinded to one another's judgments and to the judgments of the CEC Agent. The reference label for each criterion was the majority vote of the three reviewers. The CEC Agent judged the same criteria from the same inputs.

Agreement was summarized with the macro-F1, accuracy and Cohen's $\kappa$\cite{cohen1960kappa}. For each class $c$ (met or not met), with $\mathrm{TP}_c$, $\mathrm{FP}_c$ and $\mathrm{FN}_c$ denoting the numbers of true-positive, false-positive and false-negative judgments,
\begin{equation}
\mathrm{F1}_{c}=\frac{2\,\mathrm{TP}_{c}}{2\,\mathrm{TP}_{c}+\mathrm{FP}_{c}+\mathrm{FN}_{c}},\qquad
\text{macro-F1}=\tfrac{1}{2}\left(\mathrm{F1}_{\mathrm{met}}+\mathrm{F1}_{\text{not met}}\right).
\label{eq:f1}
\end{equation}
The macro-F1 was the primary measure because criteria that were met were less common than criteria that were not met. Accuracy was the proportion of judgments that agreed with the reference label. For comparison with physician performance, each reviewer was assessed against a reference derived from the other two reviewers: their shared judgment was used when they agreed, and a fourth physician adjudicated the criterion when they disagreed. The mean across the three reviewers was then calculated. Agreement of the CEC Agent was considered acceptable if its macro-F1 was not lower than the lower quartile of the macro-F1 values of the reviewers or if its 95\% CI included the mean macro-F1 of the reviewers.

\subsection*{Analysis of sensitivity to response length of open-ended tasks}\label{sec:m13}
The analysis drew on the estimation of the relationship between score and length across conditions of verbosity in HealthBench Professional\cite{hicks2026healthbenchpro}, on the description of verbosity bias in LLM-based judges\cite{zheng2023judging} and on regression-based control for length\cite{dubois2024length}. Length adjustment was applied only to the two open-ended levels, standardized clinical scenarios and real-world clinical scenarios. 

The slope $\beta$ was estimated on one-third of the items at each open-ended level, sampled at random: 284 of the 852 standardized clinical scenarios and 122 of the 365 real-world clinical scenarios. For model $m$ and item $i$, let $L_{mi}$ denote the length of the response in characters. The item score was the positive points earned divided by the positive points available, and was zero if any safety-critical error was triggered:
\begin{equation}
S_{mi}=\mathbf{1}[E_{mi}=0]\,\frac{P_{mi}}{D_{i}}.
\label{eq:itemscore}
\end{equation}

To induce variation in length without imposing a fixed number of characters, these subsets were presented to DeepSeek-V4-Pro and Qwen3.8-Max with three prefixes to the prompt, identical at both levels: `Please interpret the report concisely but completely', `Please interpret the report' and `Please provide a detailed interpretation of the report'. The content of the input, the rubric and the inference parameters were otherwise unchanged. For model $m$ and level $g$, and for each condition $v$, we calculated the ETS $\bar{S}_{mgv}$ and the mean response length in characters $\bar{L}_{mgv}$, and fitted by least squares
\begin{equation}
\bar{S}_{mgv}=\alpha_{mg}+\beta_{mg}\bar{L}_{mgv}+\varepsilon_{mgv}
\label{eq:slope}
\end{equation}
to the three conditions, yielding slopes of $7.928 \times 10^{-5}$ (95\% CI, $5.556 \times 10^{-5}$ to $1.035 \times 10^{-4}$) and $4.202 \times 10^{-5}$ (95\% CI, $2.388 \times 10^{-5}$ to $6.022 \times 10^{-5}$) per character for DeepSeek-V4-Pro and Qwen3.8-Max on standardized clinical scenarios, and $5.179 \times 10^{-5}$ (95\% CI, $3.250 \times 10^{-5}$ to $7.141 \times 10^{-5}$) and $5.710 \times 10^{-5}$ (95\% CI, $3.333 \times 10^{-5}$ to $8.321 \times 10^{-5}$) on real-world clinical scenarios. The 95\% CIs were the 2.5th and 97.5th percentiles of 5,000 replicates that drew cases with replacement within the estimation subset of that level (284 and 122); each drawn case kept both models under the three prompts, with criterion-level scores and lengths, and the ETS, mean length and slope were recalculated.

The coefficient of adjustment weighted the two levels equally. The two model slopes were averaged within each level without weighting, and the two level means were then averaged without weighting and truncated at zero,
\begin{equation}
\beta_{g}=\frac{1}{2}\sum_{m}\beta_{mg},\qquad
\beta=\max\!\left(0,\ \frac{1}{2}\left(\beta_{\mathrm{std}}+\beta_{\mathrm{real}}\right)\right),
\label{eq:beta}
\end{equation}
which equalled $5.75 \times 10^{-5}$ per character, corresponding to an increase in the ETS of 0.058 per 1,000 characters. The same coefficient was applied to standardized clinical scenarios and to real-world clinical scenarios.

The reference length $L_{0}$ was the zero point of the adjustment: a response whose length equalled $L_{0}$ was neither penalized nor credited. For each level, $L_{0}$ was the median length, in characters, of the default responses of all ten models to every item at that level: 1,151 for standardized clinical scenarios and 1,730 for real-world clinical scenarios. Each response was adjusted separately, retaining the item-level safety gate and a lower bound of zero:
\begin{equation}
S_{mi}^{\mathrm{adj}}=\mathbf{1}[E_{mi}=0]\,\max\!\left(0,\;S_{mi}-\beta\,(L_{mi}-L_{0})\right).
\label{eq:adj}
\end{equation}
Responses with any safety-critical error remained at zero; otherwise, the adjustment reduced the score of a longer response, subject to the lower bound, and increased that of a shorter response. No upper bound was imposed, so adjusted scores and their weighted aggregate could exceed 1. For the items $G$ at one open-ended level, the length-adjusted ETS was
\begin{equation}
\mathrm{ETS}^{\mathrm{adj}}_{m}
=\frac{\sum_{i\in G} D_{i}\,S_{mi}^{\mathrm{adj}}}{\sum_{i\in G} D_{i}}.
\label{eq:etsadj}
\end{equation}
The length-adjusted ETS was reported on all open-ended items, using the original responses of the ten models to the 852 standardized and 365 real-world clinical scenarios, and was compared with the unadjusted ETS.

\subsection*{Statistical analysis}\label{sec:m14}
Analyses were descriptive. Scores are reported as point estimates with 95\% percentile CIs obtained from 5,000 case-level bootstrap resamples. All criteria and model outputs belonging to an item were resampled together, and scores were recalculated in each resample. Resampling was stratified by evaluation level for pooled estimates and conducted within each displayed subgroup for estimates by clinical intent or department. For guideline-knowledge items, 95\% CIs of the proportion of items answered correctly were calculated with the Wilson score method\cite{wilson1927probable}. The concordance of model rankings between levels was summarized with Spearman's rank correlation coefficient, with mid-ranks assigned to ties; 95\% CIs were obtained by Fisher transformation with the variance estimator of Bonett and Wright\cite{bonett2000sample} ($n = 10$ models). Summary statistics across models are the unweighted mean and range of the ten model-level scores unless stated otherwise. For the agreement statistics, 95\% CIs were obtained from 5,000 cluster bootstrap resamples of the 159 items\cite{efron1993bootstrap}, retaining all judgments of each drawn item and recalculating the statistics overall and by stage within each resample; the physician mean was the unweighted mean of the three physicians in that resample. Differences between models were not tested formally and no adjustment was made for multiple comparisons; comparisons between models, and in particular those within subgroups, are therefore descriptive. No statistical method was used to predetermine the number of items. Analyses were performed with Python 3.14.3 (Anaconda, Inc.; GCC 14.3.0) on Linux (kernel 6.8.0-138-generic, x86\_64, glibc 2.35).

\subsection*{Use of large language models}\label{sec:m15}
LLMs were the object of this evaluation and were components of the MEB-QC Agent and the CEC Agent, as described above.

\section*{Data availability}
The public evaluation interface for Med Evidence Bench, including the leaderboard and a description of the benchmark, is available at \url{https://www.medevidencebench.org.cn/}.

\bibliography{meb-bibliography}

\section*{Competing interests}
The authors declare no competing interests.

\section*{Additional information}
\textbf{Extended data} is available for this paper (Extended Data Figs.~1 and~2 and Extended Data Tables~1--4).

\FloatBarrier
\clearpage
\nolinenumbers  
\newgeometry{left=2cm,right=2cm,top=2cm,bottom=2.2cm}
\iffiguresinline\else
\section*{Figures}

\figI{H}{\textwidth}
\clearpage
\figII{H}{0.80\textwidth}
\clearpage
\figIII{H}{\textwidth}
\clearpage
\figIV{H}{\textwidth}
\clearpage
\figV{H}{\textwidth}
\clearpage
\figVI{H}{\textwidth}
\clearpage
\fi


\begin{landscape}
\section*{Extended Data}
\begin{figure}[H]
\centering
\includegraphics[width=\linewidth]{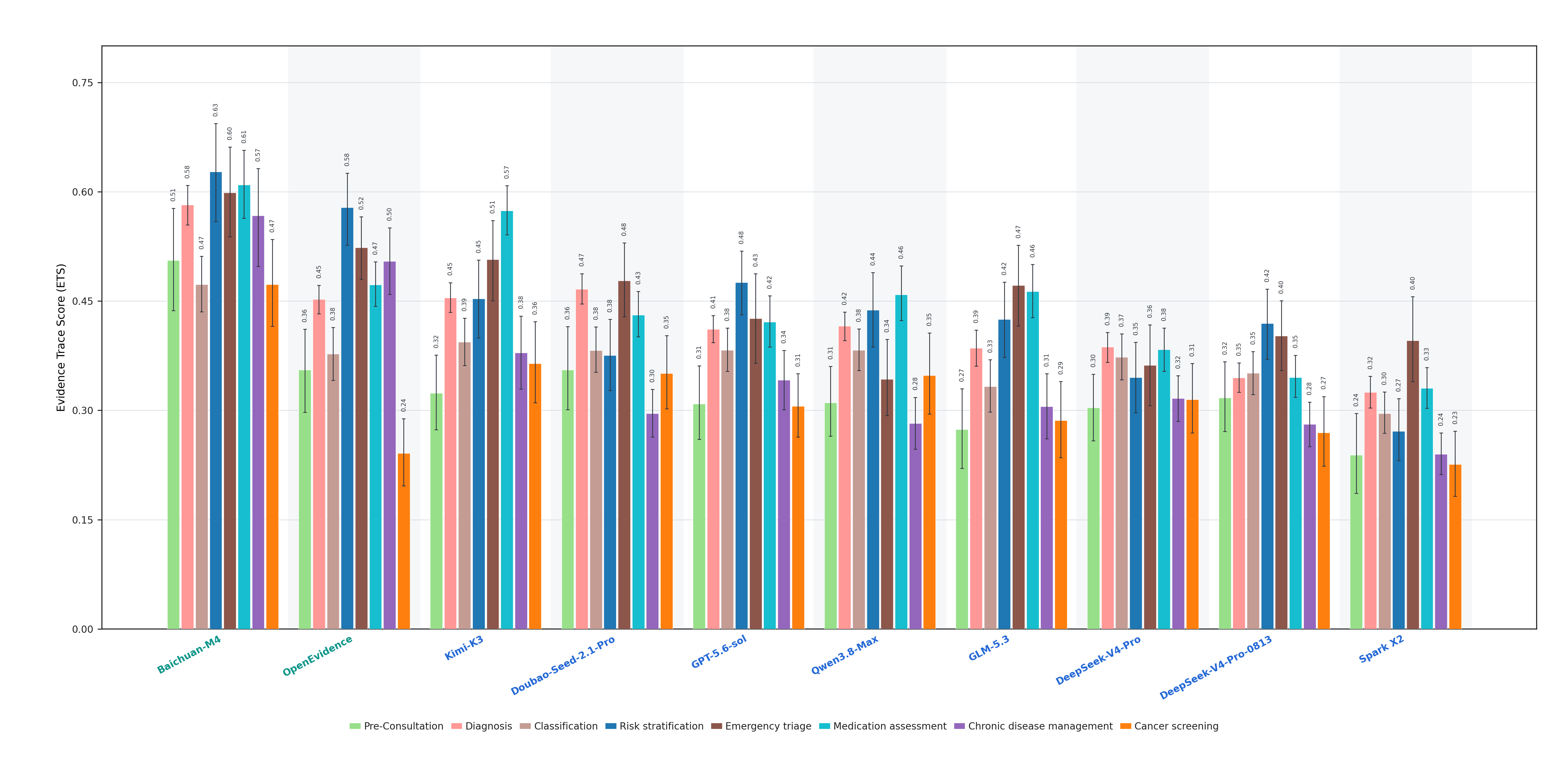}
\caption*{\textbf{Extended Data Fig. 1 $|$ ETS by clinical intent.} ETS of each model across all 1,536 items, grouped by the eight clinical intents. Values above the bars are point estimates; error bars show 95\% CIs obtained from 5,000 case-level bootstrap resamples conducted within each clinical-intent group. The number of items for each intent is given in Extended Data Table~2. Model names in green denote medical systems and those in blue denote general-purpose models.}
\label{edfig:1}
\end{figure}
\end{landscape}

\begin{landscape}
\begin{figure}[H]
\centering
\includegraphics[width=\linewidth]{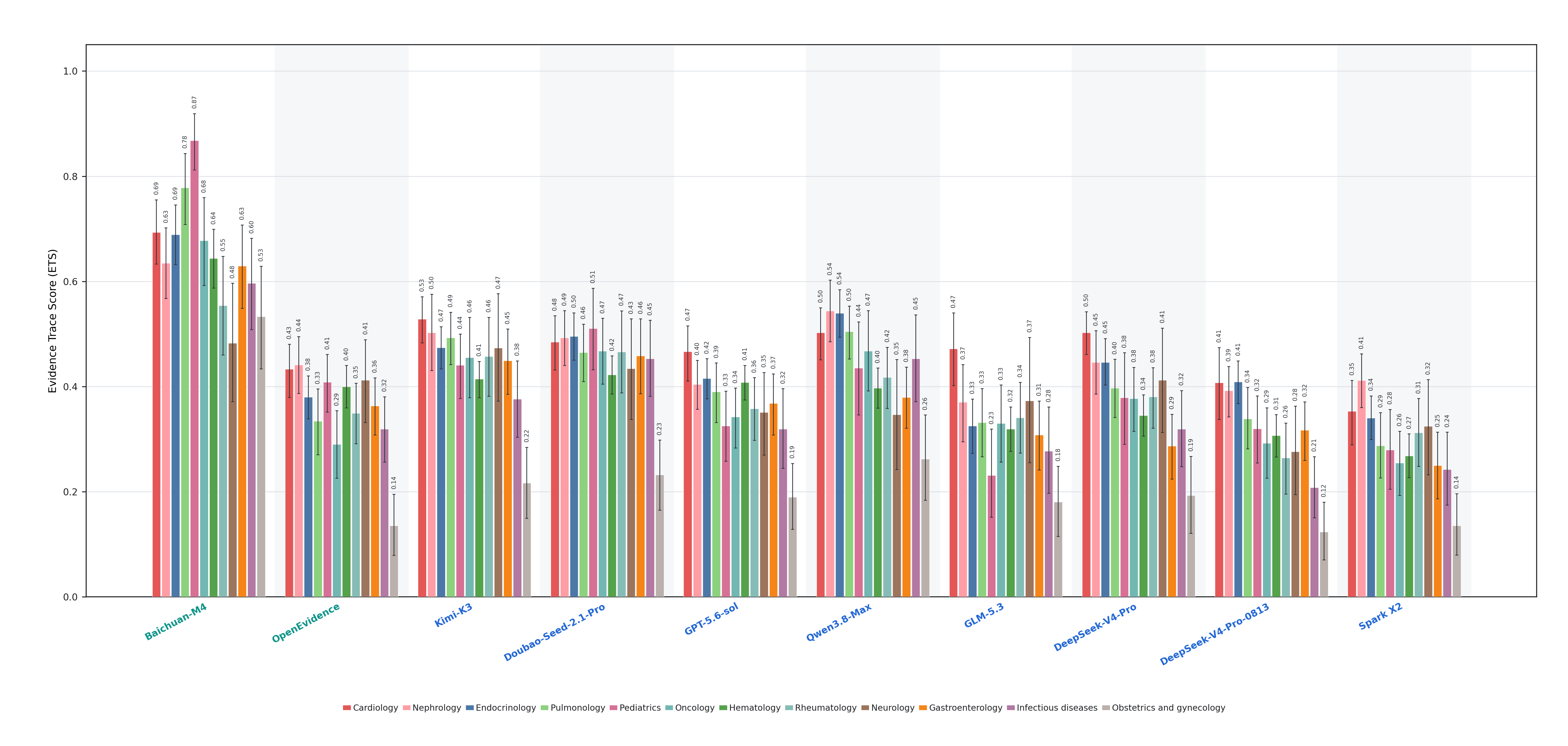}
\caption*{\textbf{Extended Data Fig. 2 $|$ ETS by clinical department in standardized clinical scenarios.} ETS of each model across the 852 standardized clinical scenarios, grouped by the 12 clinical departments. Values above the bars are point estimates; error bars show 95\% CIs obtained from 5,000 case-level bootstrap resamples conducted within each clinical department. Model names in green denote medical systems and those in blue denote general-purpose models.}
\label{edfig:2}
\end{figure}
\end{landscape}

\restoregeometry
\newgeometry{left=1.2cm,right=1.2cm,top=1.8cm,bottom=1.8cm}

\begin{landscape}
\noindent{\EDtabfont\bfseries Extended Data Table 1 $|$ Evaluated systems and inference settings}\par
\vspace{0.5em}
{\EDtabfont\footnotesize
\setlength{\tabcolsep}{3pt}%
\renewcommand{\arraystretch}{1.25}%
\noindent\begin{tabular}{@{}Y{2.6cm} Y{1.3cm} Y{2.4cm} Y{2.2cm} Y{1.5cm} Y{2.1cm} Y{2.1cm} Y{3.9cm} Y{3.8cm} Y{2.2cm}@{}}
\toprule
Model &
Maximum output tokens &
Temperature &
Top \textit{p} &
Top \textit{k} &
Frequency penalty &
Presence penalty &
Reasoning mode &
Version, release date or knowledge cut-off &
Access \\
\midrule
Baichuan-M4 & 10,000 & Not reported & Not reported & Not reported & Not reported & Not reported & Not reported & Released 22 June 2026 & Closed source (API) \\
DeepSeek-V4-Pro & 10,000 & Inactive when thinking mode enabled & 0.95 (lower bound, thinking mode) & Not reported & Inactive when thinking mode enabled & Inactive when thinking mode enabled & Thinking enabled (API default); reasoning effort high (API default) & Preview, 24 April 2026 & Open source \\
DeepSeek-V4-Pro-0813 & 10,000 & Inactive when thinking mode enabled & 0.95 (lower bound, thinking mode) & Not reported & Inactive when thinking mode enabled & Inactive when thinking mode enabled & Thinking enabled (API default); reasoning effort high (API default) & General availability, August 2026 & Open source \\
Doubao-Seed-2.1-Pro & 10,000 & 1 & 0.7 & Not reported & Not supported & Not supported & Thinking enabled (API default); reasoning effort high (API default) & 23 June 2026 & Closed source (API) \\
GLM-5.3 & 10,000 & 1 & 0.95 & Not reported & Not reported & Not reported & Thinking always on; effort maximum (API default) & Released 14 August 2026 & Closed source (API) \\
GPT-5.6-sol & 10,000 & Ignored & Ignored & Not supported & Ignored & Ignored & Thinking enabled; effort high (set) & Knowledge cut-off 16 February 2026; released 9 July 2026 & Closed source (API) \\
Kimi-K3 & 10,000 & 1.0 (fixed) & 0.95 (fixed) & Not reported & 0 (fixed) & 0 (fixed) & Thinking always on; effort maximum (API default) & Knowledge cut-off April 2026; weights released 27 July 2026 & Open source \\
OpenEvidence & 10,000 & NA & NA & NA & NA & NA & NA & Accessed May 2026 & Web interface \\
Qwen3.8-Max & 10,000 & Not reported & Not reported & 20 & Not reported & Not reported & Thinking enabled (API default); effort extra high (API default) & Released 2 August 2026; version of 2 September 2026 & Closed source (API) \\
Spark X2 & 10,000 & 1.2 & 0.95 & 6 & 0.001 & 2.01 & Thinking enabled (API default); reasoning effort not available & 11 February 2026 & Closed source (API) \\
\bottomrule
\end{tabular}%
\par\vspace{0.6em}
\noindent{\EDtabfont\footnotesize The maximum number of output tokens was set by the investigators; all other parameters were left at each provider's default values. OpenEvidence was accessed through its web interface. API, application programming interface; NA, not applicable; not reported, a value not documented by the provider. }
}
\label{edtab:1}
\end{landscape}

\noindent{\EDtabfont\bfseries Extended Data Table 2 $|$ Distribution of items and criteria by clinical intent}\par
\vspace{0.75em}
{\EDtabfont\footnotesize
\setlength{\tabcolsep}{2.6pt}%
\renewcommand{\arraystretch}{1.15}%
\noindent\begin{tabular}{@{}L
>{\EDtabfont\centering\arraybackslash}m{1.1cm}
>{\EDtabfont\centering\arraybackslash}m{1.15cm}
>{\EDtabfont\centering\arraybackslash}m{1.6cm}
>{\EDtabfont\centering\arraybackslash}m{1.9cm}
>{\EDtabfont\centering\arraybackslash}m{1.6cm}
>{\EDtabfont\centering\arraybackslash}m{1.5cm}
>{\EDtabfont\centering\arraybackslash}m{1.5cm}
>{\EDtabfont\centering\arraybackslash}m{1.2cm}
>{\EDtabfont\centering\arraybackslash}m{1.75cm}
@{}}
\toprule
 & \multicolumn{5}{c}{\EDtabfont Number of items} & \multicolumn{4}{c}{\EDtabfont Number of criteria} \\
\cmidrule(lr){2-6}\cmidrule(l){7-10}
Clinical intent &
All &
\% of all\newline items &
Guideline\newline knowledge &
Standardized\newline clinical\newline scenarios &
Real-world\newline clinical\newline scenarios &
Clinical\newline foundations &
Clinical\newline reasoning &
Clinical\newline safety &
Evidence-based\newline decision-making \\
\midrule
Diagnosis & 532 & 34.64 & 78 & 336 & 118 & 212 & 427 & 466 & 979 \\
Medication assessment & 254 & 16.54 & 72 & 164 & 18 & 90 & 98 & 218 & 418 \\
Classification & 244 & 15.89 & 66 & 104 & 74 & 146 & 197 & 252 & 448 \\
Risk stratification & 143 & 9.31 & 37 & 64 & 42 & 79 & 170 & 190 & 305 \\
Chronic disease management & 114 & 7.42 & 38 & 44 & 32 & 70 & 134 & 166 & 216 \\
Emergency triage & 96 & 6.25 & 22 & 36 & 38 & 74 & 62 & 80 & 204 \\
Cancer screening & 88 & 5.73 & 6 & 64 & 18 & 12 & 58 & 118 & 188 \\
Pre-consultation & 65 & 4.23 & 0 & 40 & 25 & 26 & 39 & 77 & 192 \\
Total & 1,536 & 100 & 319 & 852 & 365 & 709 & 1,185 & 1,567 & 2,950 \\
\bottomrule
\end{tabular}%
\par\vspace{0.6em}
\noindent{\EDtabfont\footnotesize The total number of criteria is 6,411.}
}
\label{edtab:2}

\clearpage

\noindent{\EDtabfont\bfseries Extended Data Table 3 $|$ Agreement of the automated grader and of individual physicians with reference labels}\par
\vspace{0.75em}
{\EDtabfont\scriptsize
\setlength{\tabcolsep}{2.6pt}%
\renewcommand{\arraystretch}{1.12}%
\noindent\begin{tabular}{@{}%
  >{\EDtabfont}w{c}{1.70cm}%
  >{\EDtabfont}w{c}{1.50cm}%
  >{\EDtabfont}w{c}{2.00cm}%
  >{\EDtabfont}w{c}{2.40cm}%
  >{\EDtabfont}w{c}{1.60cm}%
  >{\EDtabfont}w{c}{1.90cm}%
  >{\EDtabfont}w{c}{2.00cm}%
  >{\EDtabfont}w{c}{1.60cm}%
  C@{}}
\toprule
\bfseries Stage &
\bfseries n &
\bfseries Macro-F1 &
\bfseries 95\% CI &
\makecell[c]{\bfseries F1\\\bfseries (met)} &
\makecell[c]{\bfseries F1\\\bfseries (not met)} &
\bfseries Accuracy &
\bfseries $\kappa$ &
\makecell[c]{\bfseries Difference\\\bfseries (grader $-$ physicians)} \\
\midrule
\multicolumn{9}{@{}l}{\EDtabfont\bfseries\itshape Automated judge: CEC Agent} \\
Overall & 1,296 & 0.9302 & 0.915--0.945 & 0.8978 & 0.9626 & 0.9452 & 0.8605 & 0.0257 \\
C1 & 174 & 0.9123 & 0.845--0.966 & 0.9728 & 0.8519 & 0.9540 & 0.8246 & 0.0397 \\
C2 & 120 & 0.9684 & 0.931--1.000 & 0.9538 & 0.9829 & 0.9750 & 0.9368 & 0.0061 \\
E1 & 318 & 0.9020 & 0.849--0.947 & 0.8293 & 0.9747 & 0.9560 & 0.8040 & 0.0306 \\
E2 & 153 & 0.8484 & 0.789--0.902 & 0.8345 & 0.8623 & 0.8497 & 0.6974 & 0.0478 \\
C3 & 234 & 0.8731 & 0.811--0.925 & 0.7901 & 0.9561 & 0.9274 & 0.7476 & 0.0641 \\
C4 & 297 & 0.9064 & 0.825--0.970 & 0.8235 & 0.9893 & 0.9798 & 0.8128 & $-$0.0236 \\
\addlinespace
\multicolumn{9}{@{}l}{\EDtabfont\bfseries\itshape Physician 1} \\
Overall & 1,296 & 0.9132 & 0.894--0.931 & 0.8752 & 0.9512 & 0.9298 & 0.8263 & --- \\
C1 & 174 & 0.8581 & 0.783--0.919 & 0.9470 & 0.7692 & 0.9138 & 0.7170 & --- \\
C2 & 120 & 0.9689 & 0.917--1.000 & 0.9552 & 0.9827 & 0.9750 & 0.9379 & --- \\
E1 & 318 & 0.9040 & 0.851--0.947 & 0.8333 & 0.9746 & 0.9560 & 0.8082 & --- \\
E2 & 153 & 0.8430 & 0.782--0.897 & 0.8481 & 0.8378 & 0.8431 & 0.6860 & --- \\
C3 & 234 & 0.8195 & 0.755--0.877 & 0.7184 & 0.9205 & 0.8761 & 0.6398 & --- \\
C4 & 297 & 0.9064 & 0.829--0.967 & 0.8235 & 0.9893 & 0.9798 & 0.8129 & --- \\
\addlinespace
\multicolumn{9}{@{}l}{\EDtabfont\bfseries\itshape Physician 2} \\
Overall & 1,296 & 0.9175 & 0.901--0.932 & 0.8837 & 0.9513 & 0.9313 & 0.8353 & --- \\
C1 & 174 & 0.9266 & 0.864--0.973 & 0.9759 & 0.8772 & 0.9598 & 0.8532 & --- \\
C2 & 120 & 0.9597 & 0.917--0.991 & 0.9429 & 0.9765 & 0.9667 & 0.9194 & --- \\
E1 & 318 & 0.8924 & 0.840--0.937 & 0.8140 & 0.9709 & 0.9497 & 0.7852 & --- \\
E2 & 153 & 0.8423 & 0.780--0.898 & 0.8537 & 0.8310 & 0.8431 & 0.6855 & --- \\
C3 & 234 & 0.7983 & 0.738--0.855 & 0.7000 & 0.8966 & 0.8462 & 0.6062 & --- \\
C4 & 297 & 0.9688 & 0.920--1.000 & 0.9412 & 0.9964 & 0.9933 & 0.9376 & --- \\
\addlinespace
\multicolumn{9}{@{}l}{\EDtabfont\bfseries\itshape Physician 3} \\
Overall & 1,296 & 0.8828 & 0.862--0.902 & 0.8291 & 0.9365 & 0.9074 & 0.7667 & --- \\
C1 & 174 & 0.8331 & 0.744--0.905 & 0.9492 & 0.7170 & 0.9138 & 0.6672 & --- \\
C2 & 120 & 0.9582 & 0.917--0.990 & 0.9394 & 0.9770 & 0.9667 & 0.9165 & --- \\
E1 & 318 & 0.8177 & 0.745--0.878 & 0.6761 & 0.9593 & 0.9277 & 0.6393 & --- \\
E2 & 153 & 0.7166 & 0.643--0.784 & 0.6906 & 0.7425 & 0.7190 & 0.4750 & --- \\
C3 & 234 & 0.8092 & 0.743--0.870 & 0.7000 & 0.9185 & 0.8718 & 0.6218 & --- \\
C4 & 297 & 0.9149 & 0.829--0.979 & 0.8387 & 0.9911 & 0.9832 & 0.8299 & --- \\
\addlinespace
\multicolumn{9}{@{}l}{\EDtabfont\bfseries\itshape Mean of physicians} \\
Overall & 1,296 & 0.9045 & 0.888--0.919 & 0.8626 & 0.9463 & 0.9228 & 0.8095 & --- \\
C1 & 174 & 0.8726 & 0.813--0.920 & 0.9574 & 0.7878 & 0.9291 & 0.7458 & --- \\
C2 & 120 & 0.9623 & 0.924--0.992 & 0.9458 & 0.9787 & 0.9694 & 0.9246 & --- \\
E1 & 318 & 0.8714 & 0.820--0.913 & 0.7744 & 0.9683 & 0.9444 & 0.7442 & --- \\
E2 & 153 & 0.8006 & 0.744--0.852 & 0.7975 & 0.8038 & 0.8017 & 0.6155 & --- \\
C3 & 234 & 0.8090 & 0.755--0.857 & 0.7061 & 0.9119 & 0.8647 & 0.6226 & --- \\
C4 & 297 & 0.9300 & 0.879--0.970 & 0.8678 & 0.9923 & 0.9854 & 0.8601 & --- \\
\bottomrule
\end{tabular}%
\par\vspace{0.6em}
\noindent{\EDtabfont\footnotesize
Agreement was assessed on 159 items from the real-world subset, with three items sampled at random from each of the 53 diseases or conditions, using responses generated by DeepSeek-V4-Pro. Three senior physicians holding the rank of chief physician in China, none of whom had participated in developing the rubrics, independently judged whether each criterion was met. They were blinded to one another's judgments and to those of the CEC Agent. The CEC Agent was assessed against the majority vote of the three physicians. Each physician was assessed against a reference derived from the other two physicians: their shared judgment was used when they agreed, and a fourth physician adjudicated the criterion when they disagreed. Each evaluator was assessed on all 1,296 criterion-level judgments. Macro-F1 is the unweighted mean of the F1 values for the met and not-met classes; $\kappa$, Cohen's kappa. The 95\% CIs are percentile intervals from 5,000 cluster bootstrap resamples of the 159 items, with all judgments of a sampled item retained together; the interval for the mean of the physicians is that of the resample-specific arithmetic means of the three physicians. Difference (grader $-$ physicians) is the macro-F1 of the CEC Agent minus the mean macro-F1 of the three physicians. C1, clinical information identification; C2, clinical calculation; E1, evidence matching; E2, evidence application; C3, clinical recommendation; C4, critical-error detection; $n$, number of criterion-level judgments.}%
}
\label{edtab:3}

\clearpage

\noindent{\EDtabfont\bfseries Extended Data Table 4 $|$ Response length and length-adjusted ETS of open-ended tasks}\par
\vspace{0.75em}
{\EDtabfont\small
\setlength{\tabcolsep}{10pt}%
\renewcommand{\arraystretch}{1.2}%
\noindent\begin{tabular}{@{}L C C C@{}}
\toprule
\multicolumn{1}{c}{\EDtabfont\bfseries Model} &
\makecell[c]{\bfseries Mean response length\\\bfseries (characters)} &
\bfseries ETS &
\bfseries Length-adjusted ETS \\
\midrule
\multicolumn{4}{@{}l}{\EDtabfont\bfseries\itshape Open-ended tasks (n = 1217)} \\
Baichuan-M4 & 1261.4 & 0.5257 & 0.5308 \\
DeepSeek-V4-Pro & 1594.2 & 0.3071 & 0.2948 \\
DeepSeek-V4-Pro-0813 & 1017.7 & 0.2878 & 0.3002 \\
Doubao-Seed-2.1-Pro & 2277.1 & 0.3593 & 0.3177 \\
GLM-5.3 & 1045.1 & 0.3245 & 0.3407 \\
GPT-5.6-sol & 889.2 & 0.3454 & 0.3713 \\
Kimi-K3 & 1087.7 & 0.3968 & 0.4105 \\
OpenEvidence & 1694.9 & 0.402 & 0.3853 \\
Qwen3.8-Max & 1776.4 & 0.3382 & 0.3191 \\
Spark X2 & 1399.7 & 0.2404 & 0.2421 \\
\addlinespace
\multicolumn{4}{@{}l}{\EDtabfont\bfseries\itshape Standardized clinical scenarios (n = 852), $\beta=5.75 \times 10^{-5}$, $L_{0} = 1{,}151$} \\
Baichuan-M4 & 1173 & 0.6582 & 0.6579 \\
DeepSeek-V4-Pro & 1291 & 0.378 & 0.3760 \\
DeepSeek-V4-Pro-0813 & 675.5 & 0.3181 & 0.3461 \\
Doubao-Seed-2.1-Pro & 2181 & 0.453 & 0.4074 \\
GLM-5.3 & 879.9 & 0.3218 & 0.3382 \\
GPT-5.6-sol & 748.8 & 0.3739 & 0.3978 \\
Kimi-K3 & 933.1 & 0.4424 & 0.4560 \\
OpenEvidence & 1488 & 0.363 & 0.3515 \\
Qwen3.8-Max & 1655 & 0.448 & 0.4257 \\
Spark X2 & 1320 & 0.2915 & 0.2881 \\
\addlinespace
\multicolumn{4}{@{}l}{\EDtabfont\bfseries\itshape Real-world clinical scenarios (n = 365), $\beta=5.75 \times 10^{-5}$, $L_{0} = 1{,}730$} \\
Baichuan-M4 & 1467 & 0.3741 & 0.3855 \\
DeepSeek-V4-Pro & 2301 & 0.226 & 0.2019 \\
DeepSeek-V4-Pro-0813 & 1816 & 0.2531 & 0.2477 \\
Doubao-Seed-2.1-Pro & 2503 & 0.2521 & 0.2150 \\
GLM-5.3 & 1431 & 0.3275 & 0.3435 \\
GPT-5.6-sol & 1217 & 0.3128 & 0.3410 \\
Kimi-K3 & 1448 & 0.3447 & 0.3584 \\
OpenEvidence & 2178 & 0.4467 & 0.4240 \\
Qwen3.8-Max & 2059 & 0.2126 & 0.1972 \\
Spark X2 & 1586 & 0.1819 & 0.1894 \\
\bottomrule
\end{tabular}%
\par\vspace{0.6em}
\noindent{\EDtabfont\footnotesize
Response length is the mean number of characters across items at that level. Length adjustment was applied only to these open-ended tasks. For each response, the item score was the positive points earned divided by the positive points available, and was zero if any safety-critical error was triggered. The adjusted score was $\mathbf{1}[E_{mi}=0]\,\max(0,\, S_{mi}-\beta(L_{mi}-L_{0}))$, where $L_{mi}$ is the length of that response. Responses with any safety-critical error remained at zero. No upper bound was imposed on adjusted scores or their weighted aggregate. $\beta$ was $5.75 \times 10^{-5}$ per character, the unweighted mean of four least-squares slopes---those of DeepSeek-V4-Pro and Qwen3.8-Max on standardized and on real-world clinical scenarios---truncated at zero and weighting the two levels equally; the slopes were estimated on 284 standardized and 122 real-world clinical scenarios, and the same coefficient was used at both levels. $L_{0}$ was the median length of the default responses of all ten models at that level: 1,151 characters for standardized clinical scenarios and 1,730 characters for real-world clinical scenarios. The length-adjusted ETS is the mean of the adjusted item scores weighted by the positive points available.}
}
\label{edtab:4}

\restoregeometry

\end{document}